\documentclass[10pt,journal,compsoc]{IEEEtran}

\usepackage{graphicx}
\usepackage{amsmath}
\usepackage{amssymb}
\usepackage{algorithm}
\usepackage{graphics}
\usepackage{cite}
\usepackage{color}
\usepackage{ragged2e}
\usepackage{graphicx}
\usepackage{multirow}
\usepackage{booktabs}
\usepackage{makecell}
\usepackage{colortbl}
\usepackage[dvipsnames,table,xcdraw]{xcolor}
\usepackage{tikz}
\usepackage{pifont}
\usepackage[breaklinks=true,colorlinks, citecolor=blue]{hyperref}
\usepackage{microtype}
\usepackage{cleveref}

\RequirePackage{silence}
\graphicspath{{./Imgs/},{./Imgs/authors/},{./imgs/Visualization/ade20k/},{./imgs/Visualization/cityscapes/}}

\DeclareGraphicsExtensions{.pdf,.jpg,.png}

\newcommand{\addFig}[1]{}
\newcommand{\addFigs}[1]{}

\definecolor{pltblue}{RGB}{174, 199, 232}
\definecolor{pltorange}{RGB}{255, 229, 204}
\definecolor{pltgreen}{RGB}{204, 229, 204}
\definecolor{pltred}{RGB}{229, 204, 204}
\definecolor{pltpurple}{RGB}{239, 218, 230}

\definecolor{tabblue}{HTML}{1f77b4}
\definecolor{taborange}{HTML}{ff7f0e}
\definecolor{tabgreen}{HTML}{2ca02c}
\definecolor{tabred}{HTML}{d62728}
\definecolor{tabpurple}{HTML}{9467bd}
\definecolor{tabpink}{HTML}{ff0080}

\definecolor{cblue}{RGB}{173, 201, 233}
\definecolor{clblue}{RGB}{222, 234, 246}
\definecolor{corange}{RGB}{255, 152, 67}
\definecolor{lorgange}{RGB}{255, 221, 149}
\definecolor{tablegray}{RGB}{180, 180, 180}

\begin{document}

\title{Annotation-Free Open-Vocabulary Segmentation for Remote-Sensing Images}

\author{Kaiyu Li, Xiangyong Cao$^{\dag}$, Ruixun Liu, Shihong Wang, Zixuan Jiang, Zhi Wang, Deyu Meng
\IEEEcompsocitemizethanks{%
\IEEEcompsocthanksitem Kaiyu Li and Zhi Wang are with School of Software Engineering, Xi’an Jiaotong University, Xi’an, 710049, China. (E-mail: likyoo.ai@gmail.com, zhiwang@xjtu.edu.cn)
\IEEEcompsocthanksitem Xiangyong Cao and Shihong Wang are with School of Computer Science and Technology and Ministry of Education Key Laboratory of Intelligent Networks and Network Security, Xi’an Jiaotong University, Xi’an, 710049, China. (E-mail: caoxiangyong@mail.xjtu.edu.cn, jack3shihong@gmail.com)
\IEEEcompsocthanksitem Ruixun Liu is with School of Automation, Xi’an Jiaotong University, Xi’an, 710049, China.
(E-mail: liuruixun6343@gmail.com)
\IEEEcompsocthanksitem Zixuan Jiang is with College of Artificial Intelligence, Xi’an Jiaotong University, Xi’an, 710049, China. (E-mail: andrewjiang@stu.xjtu.edu.cn)
\IEEEcompsocthanksitem Deyu Meng is with School of Mathematics and Statistics and Ministry of Education Key Laboratory of Intelligent Networks and Network Security, Xi’an Jiaotong University, Xi’an, 710049, China. (E-mail: dymeng@mail.xjtu.edu.cn)
\IEEEcompsocthanksitem Xiangyong Cao is the corresponding author. %
}}

\IEEEtitleabstractindextext{%
\begin{abstract} 
\justifying{Semantic segmentation of remote sensing images is pivotal for comprehensive Earth observation, but the demand for interpreting new object categories, coupled with the high expense of manual annotation, poses significant challenges. Although open-vocabulary semantic segmentation (OVSS) offers a promising solution, existing frameworks designed for natural images are insufficient for the unique complexities of remote sensing data. They struggle with vast scale variations and fine-grained details, and their adaptation often relies on extensive, costly annotations. To address this critical gap, this paper introduces SegEarth-OV, the first framework for annotation-free open-vocabulary segmentation of remote sensing images. Specifically, we propose SimFeatUp, a universal upsampler that robustly restores high-resolution spatial details from coarse Vision-Language Model (VLM) features, correcting distorted target shapes without any task-specific post-training. We also present a simple yet effective Global Bias Alleviation operation to subtract the inherent global context from patch features, significantly enhancing local semantic fidelity. These components empower SegEarth-OV to effectively harness the rich semantics of pre-trained VLMs, making OVSS possible in optical remote sensing contexts. Furthermore, to extend the framework's universality to other challenging remote sensing modalities like Synthetic Aperture Radar (SAR) images, where large-scale pre-trained VLMs (e.g. SAR-CLIP) are unavailable and prohibitively expensive to create, we introduce AlignEarth, which is a distillation-based strategy and can efficiently transfer semantic knowledge from an optical VLM encoder to an SAR encoder, bypassing the need to build SAR foundation models from scratch and enabling universal OVSS across diverse sensor types. Extensive experiments on both optical and SAR datasets validate that our proposed SegEarth-OV can achieve dramatic improvements over the state-of-the-art methods, establishing a robust foundation for annotation-free and open-world Earth observation. All codes and models will be released at \url{https://github.com/earth-insights/SegEarth-OV-2}.
}
\end{abstract}
}

\maketitle

\IEEEdisplaynontitleabstractindextext
\IEEEpeerreviewmaketitle

\section{Introduction}
\label{sec:intro}

% Remote sensing image analysis plays an increasingly pivotal role in understanding and monitoring our planet, offering unparalleled insights into environmental changes, urban development, disaster response, and resource management~\cite{li2024learning, wang2025hypersigma, wu2023fully}.
Remote sensing image analysis has become increasingly pivotal for planetary understanding and monitoring, providing unparalleled insights into environmental dynamics, urban expansion, disaster mitigation, and sustainable resource management~\cite{li2024learning, wang2025hypersigma, wu2023fully}.
Traditionally, semantic segmentation of remote sensing images predominantly depends on supervised learning approaches, which require large-scale datasets with pixel-level annotations ~\cite{zheng2023farseg++, li2025segearthr1}. 
Despite significant advancements, this paradigm is subject to inherent limitations: the prohibitive cost and time investment for dense annotation, and its inability to recognise novel or unseen object categories in dynamic real-world scenarios.
This bottleneck has catalysed a paradigm shift towards open-vocabulary semantic segmentation (OVSS), an emerging setting that enables models to segment arbitrary object categories through textual descriptions, thereby fundamentally resolving the annotation constraint.

\begin{figure}[t]
  \centering
%   \fbox{\rule{0pt}{2in} \rule{0.9\linewidth}{0pt}}
   \includegraphics[width=1.0\linewidth]{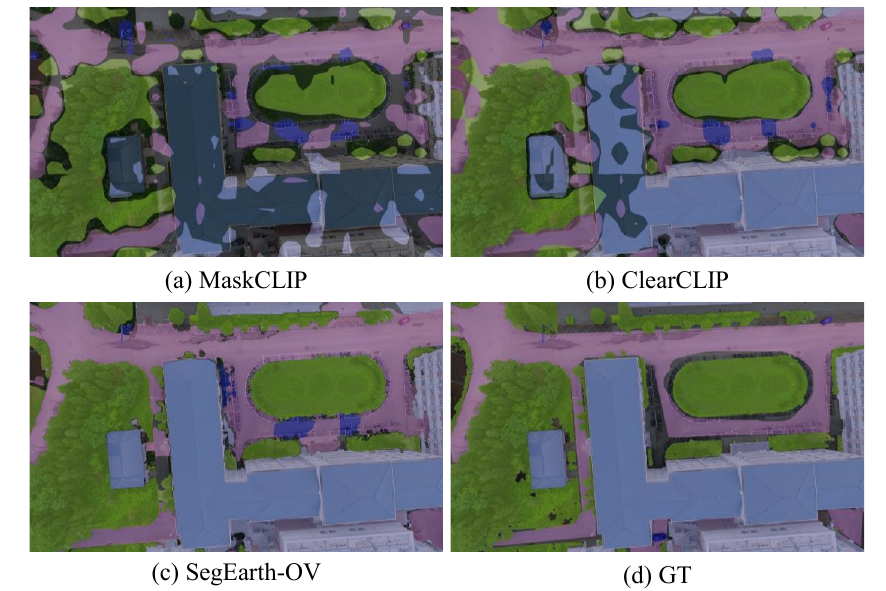}
   \caption{Limitations of state-of-the-art OVSS methods in remote sensing images. The two predictions (a) and (b) present distorted target shapes and ill-fitting boundaries (best viewed digitally with zoom, especially for the object edges).}
   \vspace{-1em}
   \label{fig:motivation}
\end{figure}

% However, adapting generic OVSS frameworks, particularly those based on pre-trained Vision-Language Models (VLMs) like CLIP~\cite{radford2021learning}, to the unique characteristics of remote sensing images presents substantial challenges.
However, the adaptation of generic OVSS frameworks, especially those leveraging pre-trained Vision-Language Models (VLMs) such as CLIP~\cite{radford2021learning}, to remote sensing images encounters substantial challenges.
Remote sensing images inherently exhibit multi-scale characteristics, encompassing objects spanning orders of magnitude, from vast land cover to fine structures such as small buildings or vehicles.
% Remote sensing images often contain objects across a vast range of scales, from expansive land cover to minute structures like small buildings or vehicles. 
The direct application of VLMs, particularly those employing aggressive downsampling strategies (\textit{e.g.}, 1/16th of the original image resolution for ViT-based CLIP), inevitably leads to irreversible loss of fine-grained spatial details.
% Directly applying VLMs, which frequently operate on downsampled features (\textit{e.g.}, 1/16th of the original image resolution for ViT-based CLIP), inevitably leads to significant loss of fine-grained spatial information. 
As shown in Fig.~\ref{fig:motivation} (a)(b), the resulting distorted target shapes and imprecise boundaries compromise the segmentation accuracy, thus undermining the effectiveness of dense prediction.
% This results in distorted target shapes, imprecise boundaries, and diminished accuracy for small objects, hindering effective dense prediction, as shown in Fig.~\ref{fig:motivation} (a)(b). 
Moreover, VLMs like CLIP are pre-trained on large-scale image-text pairs of natural images, where their [CLS] tokens encode ``global bias''~\cite{mukhoti2023open, ranasinghe2023perceptual, wang2023sclip, simeoni2025dinov3}.
% Furthermore, VLMs like CLIP are pre-trained on large-scale image-text pairs primarily from natural images, encoding a ``global bias'' through their [CLS] tokens. 
While this bias aids image-level classification, it inadvertently ``leaks'' into local patch features, negatively impacting the precise pixel-level understanding, which is essential for semantic segmentation.
% While beneficial for image-level classification, this global information can inadvertently ``contaminate'' local patch features, negatively impacting the precise pixel-level understanding required for semantic segmentation.

% To address these critical challenges, we proposed SegEarth-OV~\cite{li2025segearth}, the first annotation-free OVSS framework for remote sensing.
% % To effectively overcome these critical challenges, we introduced SegEarth-OV~\cite{li2025segearth}, the first annotation-free framework for remote sensing OVSS. 
% SegEarth-OV incorporates two core adaptations for remote sensing images: SimFeatUp and Global Bias Alleviation. Designed as a universal upsampler, after its initial general training, SimFeatUp operates without task-specific fine-tuning~\cite{li2025dynamicearth, pang2025special}. It meticulously reconstructs high-resolution spatial details from low-resolution VLM features through a parameterized Joint Bilateral Upsampling (JBU) mechanism, coupled with a Content Retention Network (CRN) that ensures content consistency. Complementary to this, Global Bias Alleviation is an elegant subtraction operation that effectively mitigates the global bias from VLM patch features, thereby enhancing local semantic fidelity and improving the precision of pixel-level predictions. Proven by extensive experimental analysis, SimFeatUp and Global Bias Alleviation have empowered SegEarth-OV to achieve state-of-the-art (SOTA) performance on diverse optical remote sensing benchmarks, establishing a robust foundation for annotation-free OVSS in this domain.

To address these critical challenges, we introduced SegEarth-OV~\cite{li2025segearth}, the first framework designed for annotation-free OVSS of remote sensing images. To tackle the issues of distorted shapes and imprecise boundaries, we propose SimFeatUp, a universal upsampler that meticulously reconstructs high-resolution spatial details from low-resolution VLM features through a parameterized Joint Bilateral Upsampling (JBU) mechanism, coupled with a Content Retention Network (CRN) that ensures content consistency. Crucially, after a one-time general training, SimFeatUp operates without any task-specific post-training~\cite{li2025dynamicearth, pang2025special}. To enhance local semantic fidelity, we present Global Bias Alleviation, a simple yet effective subtraction operation that mitigates the global bias from VLM patch features, enabling more precise pixel-level predictions. Proven by extensive experimental analysis, SimFeatUp and Global Bias Alleviation have empowered SegEarth-OV to achieve state-of-the-art (SOTA) performance on diverse optical remote sensing benchmarks, establishing a robust foundation for annotation-free OVSS in this domain.

A comprehensive Earth observation system must handle diverse data modalities beyond optical images. Synthetic Aperture Radar (SAR) images, for instance, offer all-weather, day-and-night operational capabilities, but extending open-vocabulary methods to them is non-trivial~\cite{shu2025earthmind}. The primary reason is the sheer difficulty of creating the necessary training data. Unlike optical images, SAR images are not visually intuitive; their content is determined by surface backscatter properties, speckle noise, and geometric effects that require expert knowledge to interpret correctly. Consequently, it is extremely difficult and costly to write the vast number of accurate, descriptive text captions needed for pre-training. To solve this data scarcity issue, we introduce AlignEarth. Our strategy offers a pragmatic alternative: instead of attempting the difficult task of aligning SAR images with text, AlignEarth uses knowledge distillation to transfer semantic understanding from a pre-trained optical VLM to an SAR image encoder. It achieves this by using readily available, paired optical-SAR images, even when these pairs exhibit imperfect spatial alignment or temporal discrepancies, completely bypassing the need for any SAR-specific text annotations. This design makes the trained SAR encoder fully compatible with our SimFeatUp and Global Bias Alleviation modules, thus establishing SegEarth-OV as a versatile framework for diverse remote sensing modalities.

Our main contributions can be summarized as follows:

\begin{itemize}

% 199/200行说的是SegEarth-OV's core modules，用upsampling module保持指代一致避免读者混淆
% \item We introduce SimFeatUp, a universal upsampler that requires no post-training to robustly restore lost spatial information from low-resolution VLM features, addressing distorted target shapes and imprecise boundaries in remote sensing images.
\item We introduce SegEarth-OV, the first annotation-free open-vocabulary segmentation framework specifically designed for remote sensing images, achieving SOTA performance on various remote sensing segmentation tasks.

\item We introduce SimFeatUp, a universal feature upsampling module that robustly restores the lost spatial information from low-resolution VLM features with no post-training requirements, rectifying the distorted target shapes and imprecise boundaries in remote sensing images.

% \item We propose Global Bias Alleviation, a simple yet effective subtraction operation that critically mitigates the inherent global bias from VLM patch features, thereby significantly enhancing local semantic fidelity for pixel-level predictions in remote sensing OVSS.
% fidelity 和 discriminability 
\item  We propose Global Bias Alleviation, a simple yet effective operation that subtracts the inherent global bias from VLM patch features, thereby significantly enhancing local semantic discriminability for pixel-level predictions.

% \item We introduce AlignEarth, an efficient knowledge distillation strategy that leverages combined contrastive and similarity losses on paired optical-SAR images to transfer rich semantic knowledge from optical VLMs to a SAR image encoder. This enables SegEarth-OV to perform OVSS on SAR images for the first time, significantly extending its applicability to heterogeneous remote sensing modalities.
\item We design AlignEarth, a novel knowledge distillation strategy that transfers rich semantic knowledge from optical VLM to a dedicated SAR encoder, bridging the modality gap and unlocking annotation-free OVSS for SAR images.

% \item We conduct comprehensive experiments on a wide range of diverse remote sensing datasets, covering both optical and SAR modalities. Through detailed multi-dimensional ablation studies and performance evaluations, we demonstrate that the SegEarth-OV framework exhibits outstanding performance in various remote sensing segmentation tasks.
% \item  We conduct extensive experiments on a wide range of diverse remote sensing datasets, spanning from optical to SAR modalities. Our multi-dimensional evaluation demonstrates SegEarth-OV's state-of-the-art performance on various remote sensing segmentation tasks.

\end{itemize}

The remainder of this paper is organized as follows: Section~\ref{sec:related_work} reviews related work in VLM, semantic segmentation and OVSS. Section~\ref{sec:seg_earth_ov_full} and Section~\ref{sec:alignearth} detail the core methodology of SegEarth-OV and AlignEarth. Section~\ref{sec:exps} presents extensive experimental results on diverse remote sensing datasets. Finally, Section~\ref{sec:conclusion} concludes the paper and discusses future research directions.

\section{Related Work}
\label{sec:related_work}

\subsection{Vision-Language Model}

% Recently, foundation models, especially VLMs, have energized the field of computer vision. One phenomenal advance is contrastive language-vision pretraining, \textit{i.e.}, CLIP~\cite{radford2021learning}, which elegantly bridges the gap between images and natural language.
Recent advances in foundation models, particularly VLMs, have energized the field of computer vision, among which the phenomenal contrastive language-vision pretraining, \textit{i.e.}, CLIP~\cite{radford2021learning} bridges the gap between images and natural language.
% Specifically, CLIP learns visual representations from images and aligns them with textual representations from corresponding captions in a shared embedding space. This is achieved through a contrastive learning objective, maximizing the similarity between correct image-text pairs while pushing away incorrect ones. The resulting joint embedding space allows for powerful zero-shot transfer capabilities, meaning CLIP can classify or recognize objects it has not explicitly seen during training, merely by matching visual features to new textual descriptions.
More specifically, CLIP establishes a shared embedding space through a contrastive learning objective where image representations and text representations of the corresponding captions are jointly optimized by enforcing modality alignment on matching pairs while maintaining separation between non-matching pairs. The resulting joint embedding space enables powerful zero-shot transfer, allowing CLIP to classify unseen objects by simply aligning visual features with novel textual descriptions.
This remarkable capability is what makes open-vocabulary learning possible across various downstream tasks \cite{wu2024towards, zhu2024survey}. Subsequently, related research has gradually emerged, exploring various aspects from the data collection and curation strategies \cite{cherti2023reproducible, yang2023alip, xu2023metaclip, singh2024synthetic}, efficient pre-training methodologies \cite{li2023scaling, yang2023alip, fan2024improving}, to advancements in model architectures themselves \cite{li2022blip, li2023blip}. However, CLIP, despite its impressive zero-shot abilities, primarily focuses on global \texttt{[CLS]} tokens for image-level representation. Even though patch-level tokens can be generated, they are inevitably contaminated by this global bias \cite{mukhoti2023open, ranasinghe2023perceptual, wang2023sclip}. This global semantic information, while useful for coarse classification, is often detrimental to dense prediction tasks which demand fine-grained local fidelity. For instance, the global context might incorrectly activate irrelevant local regions, leading to imprecise segmentation. 

In the context of remote sensing, specialized VLMs have also emerged, adapting general VLMs or developing new architectures to handle the unique characteristics of aerial and satellite image. For example, RemoteCLIP~\cite{liu2024remoteclip}, RS5M~\cite{zhang2023rs5m} and SkyCLIP~\cite{wang2024skyscript} adapt CLIP to remote sensing by training it on a large dataset of remote sensing image-text pairs, while H2RSVLM~\cite{pang2024h2rsvlm} focuses on building helpful and honest remote sensing VLMs to address potential biases. These works aim to bridge the domain gap between natural and remote sensing images, and extract more meaningful features for Earth observation tasks~\cite{hong2024spectralgpt, pang2024hsigene}. Building upon these efforts, our proposed SegEarth-OV extends open vocabulary interpretation to the pixel level; our proposed AlignEarth further distills an image encoder customized for remote sensing images beyond optical images, with the potential to achieve open vocabulary interpretation for full-spectrum remote sensing.

% Furthermore, as the remote sensing domain is highly specialized, several remote sensing VLMs have emerged to adapt general VLMs to remote sensing contexts \cite{liu2024remoteclip, zhang2023rs5m, wang2024skyscript, pang2024h2rsvlm} or to explicitly mine the unique characteristics of remote sensing data~\cite{hong2024spectralgpt, pang2024hsigene}. For example, RemoteCLIP~\cite{liu2024remoteclip} and RS5M~\cite{zhang2023rs5m} adapt CLIP to remote sensing by training it on a large dataset of remote sensing image-text pairs, while H2RSVLM~\cite{pang2024h2rsvlm} focuses on building helpful and honest remote sensing VLMs to address potential biases.

% These domain-specific VLMs improve performance within the optical remote sensing domain but generally do not inherently address the cross-modal challenges posed by SAR imagery, which is a key focus of our work.

\subsection{Semantic Segmentation with Upsampler}

Semantic segmentation aims to discriminate images at the pixel level. The prediction head (aka decoder), as an essential component of segmentation models, is able to upsample low-resolution feature maps into high-resolution predictions. Typical prediction heads leverage upsampling operators (\textit{e.g.}, bilinear interpolation, JBU \cite{kopf2007joint}) and high-resolution encoder features (as guidance), \textit{e.g.}, UNet \cite{ronneberger2015u}, UperNet \cite{xiao2018unified}, Semantic FPN \cite{kirillov2019panoptic}, MaskFormer \cite{cheng2021per}, \textit{etc}. Some works \cite{liu2023learning, lu2022sapa, zhou2024refreshed} focus on dynamic, learnable upsampling operators that make this process content-aware. For instance, Liu~\textit{et.al.}~\cite{liu2023learning} reconceptualized upsampling as a point sampling problem, developing a lightweight and efficient operator that bypasses computationally expensive dynamic convolutions. Building on similarity-based paradigms, Zhou~\textit{et.al.}~\cite{zhou2024refreshed} introduced ReSFU, which systematically enhances the guidance-based upsampling pipeline, including explicit query-key alignment and a flexible similarity measure.
% , to achieve superior performance, particularly in challenging direct high-ratio upsampling scenarios.

FeatUp \cite{fu2024featup} proposes a model-agnostic framework for producing high-resolution deep features. It leverages multi-view consistency, akin to NeRF~\cite{mildenhall2021nerf}, to reconstruct lost spatial information.
% through either a fast feedforward JBU network or an implicit per-image model.
This approach significantly enhances spatial resolution and improves performance for downstream dense prediction tasks like semantic segmentation, serving as a drop-in feature replacement for existing decoders. \textbf{However, FeatUp primarily showcases its utility and benchmarks its performance within supervised or transfer learning settings, where explicit ground-truth labels are required.} Inspired by FeatUp's success in producing semantically rich, high-resolution feature representations, the SimFeatUp proposed in this work extends this powerful paradigm to significantly enhance OVSS in an entirely annotation-free manner, thus tackling the challenge of generalization to novel categories without requiring any human annotations.

% \textbf{However, it only explores the condition with labels.} Inspired by FeatUp and built on top of it, the SimFeatUp proposed in this work is able to significantly improve OVSS without any labels.

\subsection{Semantic Segmentation in Remote Sensing}

Remote sensing image semantic segmentation aims to assign a predefined semantic category (\textit{e.g.}, building, road, water body) to each pixel in an image, serving as a basic task in land cover mapping, environmental monitoring, \textit{etc}. To tackle the unique challenges of remote sensing images, such as vast scale variations and complex object distributions, numerous specialized supervised models have been proposed~\cite{zheng2020foreground, zheng2023farseg++, liu2024rotated, loveda, wang2023samrs, garioud2023flair}. For instance, FarSeg~\cite{zheng2020foreground} introduced a foreground-aware relation network to address the foreground-background imbalance and the high intra-class variance of the background, a concept further refined in FarSeg++~\cite{zheng2023farseg++}.

Despite these architectural advances, these methods operate under a supervised, closed-set paradigm, which leads to two major bottlenecks: (1) They heavily rely on large-scale, pixel-level annotated datasets. Annotating remote sensing images is not only costly and time-consuming but also requires professional geographical expertise, severely limiting the scalability and applicability of the models. (2) Models trained under a closed-set assumption cannot recognize or segment new categories not present in the training set, a significant limitation in dynamic Earth scenarios. To overcome these constraints, researchers have begun exploring new paradigms such as weakly-supervised~\cite{li2024learning, zhang2024weakly}, few-shot learning~\cite{wang2024class, li2024generalized}, and the focus of this paper, OVSS, which aims to break the reliance on dense annotations and closed category sets.

\subsection{Open-Vocabulary Semantic Segmentation}
\label{sec:related_work_ovss}

As VLMs have shown remarkable zero-shot inference in image classification \cite{radford2021learning}, which naturally extends to semantic segmentation. They empower the segmentation pipeline to recognize seen and unseen categories, and users can segment almost any category in an image using prompt vocabulary \cite{wu2024towards, zhu2024survey}. We divide current CLIP-based OVSS methods into two groups: annotation-required and annotation-free. The former allows models to be trained on some base classes in a supervised or weakly supervised manner. Typically, some works \cite{ghiasi2022scaling, mukhoti2023open, ranasinghe2023perceptual, luo2023segclip, wuclipself} try to train a localization-aware CLIP which can naturally make dense predictions, while others \cite{li2022language, ding2022decoupling, xu2023side, xu2023san, cho2024cat, liu2024open} select a subset of the CLIP's pre-trained parameters and/or introduce a limited number of trainable parameters into the frozen CLIP, \textit{i.e.}, fine-tuning the CLIP to adapt to dense prediction on base classes.
For instance, CAT-Seg~\cite{cho2024cat} proposes a cost aggregation framework that fine-tunes CLIP's encoders by aggregating the multi-modal cosine similarity scores between image and text embeddings, thereby adapting CLIP to pixel-level tasks for both seen and unseen classes while mitigating overfitting.

Still, annotation-free OVSS methods emphasize tapping into CLIP's inherent localization capabilities with limited surgery of features or structures. MaskCLIP \cite{zhou2022extract} pioneers the removal of query and key projections at the attention pooling layer of CLIP's image encoder. Following it, subsequent studies \cite{li2023clip, wang2023sclip, bousselham2024grounding, lan2024clearclip} adequately explore self-self attention (\textit{i.e.}, \textit{q-q}, \textit{k-k} or \textit{v-v} self-attention), and these modifications somewhat mitigate noisy activations and spatial invariant perception of CLIP. Another stream \cite{shao2024explore, lavg, sun2024clip, barsellotti2024training} is the two-stage method, which first generates category-agnostic mask proposals and then classifies the masks. Besides, some other foundation models (\textit{e.g.} SAM \cite{kirillov2023segment}, Stable Diffusion \cite{rombach2021highresolution}) can be introduced to enhance the localization ability of CLIP, and these explorations also make sense \cite{lan2025proxyclip, barsellotti2024training, wang2023diffusion, li2025dynamicearth}.

Different from previous methods, our SegEarth-OV specifically addresses the inherent characteristics of remote sensing images. Contemporaneous remote sensing OVSS works~\cite{cao2024open, ye2025towards, zhu2025skysense} are annotation-required, while our framework provides an effective solution for annotation-free OVSS. Our SimFeatUp is designed as a universal upsampler; its initial training on a small dataset of images is separate from the OVSS task itself, allowing its weights to be directly applied to features from any remote sensing data for pixel-level refinement. For remote sensing images of other modalities (e.g. SAR images), our AlignEarth strategy further extends this annotation-free paradigm. AlignEarth enables a SAR image encoder to acquire semantic understanding from optical VLMs by leveraging readily available paired optical-SAR images~\cite{shermeyer2020spacenet, zhang2025multi, huang2021qxs, zhao2022comparative}. This strategy ensures that SegEarth-OV can universally perform OVSS across diverse remote sensing modalities, from optical to SAR, all in an annotation-free manner.

% Different from previous methods, we focus on the inherent characteristics of remote sensing images rather than the general attributes of natural images. The only contemporaneous work is \cite{cao2024open}, but it is annotation-required, like \cite{xu2023side, cho2024cat}. Our SimFeatUp only needs to be trained on a few images-only data beforehand, this process is independent of the semantic segmentation process and the trained weights can be used for almost any remote sensing data (like the foundation model in other works \cite{lan2025proxyclip, barsellotti2024training}). [说明AlignEarth部分也是annotation-free的]

\section{SegEarth-OV: OVSS for Remote Sensing}
\label{sec:seg_earth_ov_full}

% This section details the comprehensive SegEarth-OV, which is designed to address the unique challenges of OVSS across the remote sensing image. We begin by laying the groundwork with essential preliminaries, including CLIP~\cite{radford2021learning} and FeatUp~\cite{fu2024featup}. Subsequently, we introduce SegEarth-OV's two core modules: SimFeatUp, which meticulously restores lost spatial information, and Global Bias Alleviation, designed to mitigate CLIP's inherent global biases. 

This section details the comprehensive SegEarth-OV, which is designed to address the unique challenges of OVSS across the remote sensing image. As highlighted in Section~\ref{sec:intro}, existing VLM-based OVSS methods, primarily developed for natural images, suffer from loss of fine-grained spatial details due to aggressive downsampling and inherent global biases that degrade pixel-level accuracy in remote sensing contexts. To overcome these limitations, SegEarth-OV introduces two core modules: SimFeatUp, which meticulously restores lost spatial information from low-resolution VLM features, and Global Bias Alleviation, designed to effectively mitigate CLIP's inherent global biases. We first provide essential preliminaries on CLIP~\cite{radford2021learning} and FeatUp~\cite{fu2024featup} for context.

% Finally, we present AlignEarth, our novel cross-modal knowledge adaptation strategy, which extends SegEarth-OV's capabilities to Synthetic Aperture Radar (SAR) imagery, culminating in a unified, annotation-free OVSS solution for multi-modal remote sensing.

\subsection{Preliminaries}
\label{sec:preliminaries}

\begin{figure*}[t]
  \centering
%   \fbox{\rule{0pt}{2in} \rule{0.9\linewidth}{0pt}}
   \includegraphics[width=1\linewidth]{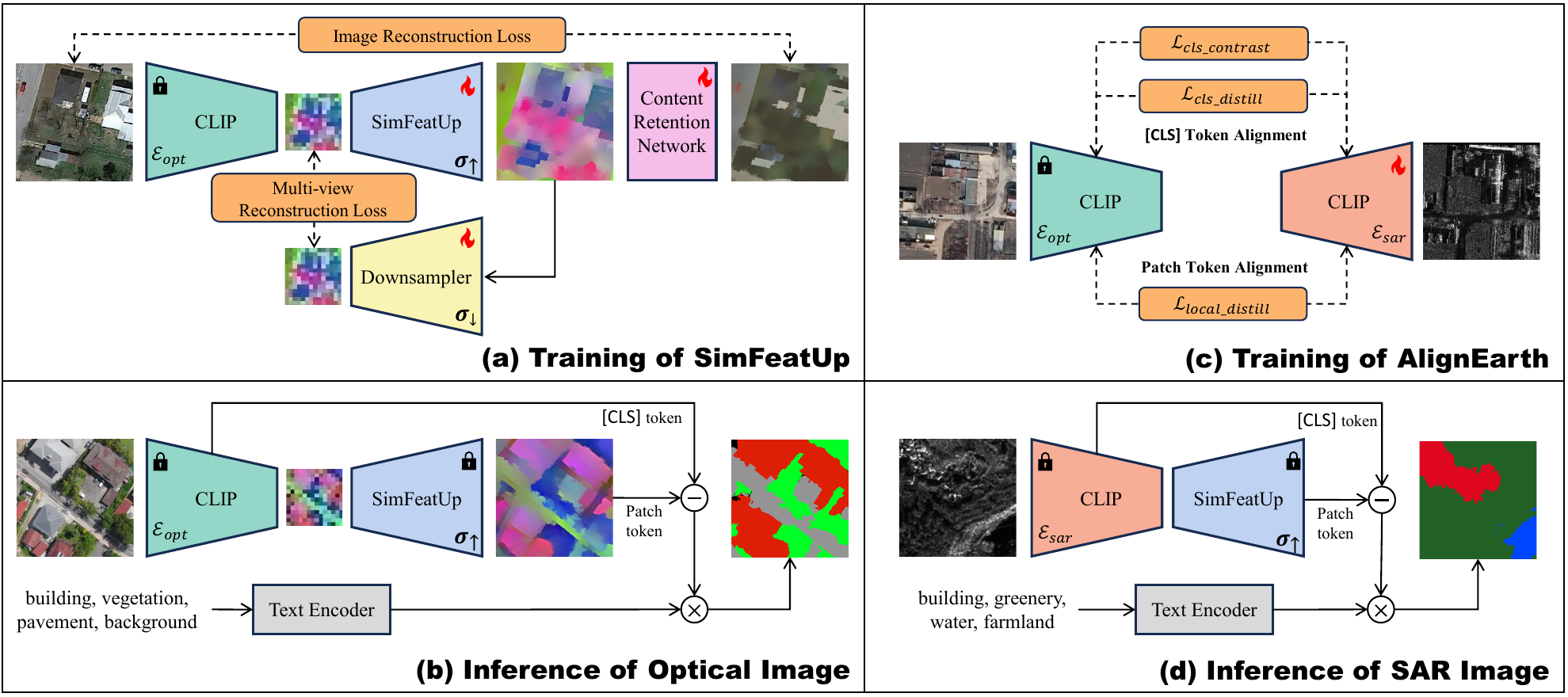}
   \caption{Illustration of the proposed method. (a) is the training process of SimFeatUp, where the CLIP image encoder is frozen and SimFeatUp's components ($\sigma_{\uparrow}$, $\sigma_{\downarrow}$, CRN) are trained using image and feature reconstruction losses. (b) is the inference process of SegEarth-OV for optical images, where low-resolution features from CLIP are upsampled by the trained SimFeatUp, and the global bias is alleviated by subtracting the [CLS] token. (c) is the training process of AlignEarth, where a trainable SAR image encoder ($\mathcal{E}_{sar}$) is aligned with a frozen optical CLIP image encoder ($\mathcal{E}_{opt}$) using contrastive and distillation losses on paired optical-SAR images. (d) is the inference process for SAR images, utilizing the AlignEarth-trained SAR image encoder, followed by SimFeatUp and global bias alleviation. For better presentation, the color renderings follow \cite{fu2024featup}.}
   \label{fig:method}
\end{figure*}

% This subsection establishes the fundamental concepts necessary for understanding our SegEarth-OV framework. We first delve into the architecture and operational principles of VLMs, particularly CLIP, which serves as the backbone for open-vocabulary capabilities. Following this, we review the principles of feature upsampling, a crucial component for dense prediction tasks that require high-resolution outputs.

\subsubsection{CLIP: A Foundation for OV Understanding}
\label{sec:clip}

VLMs have emerged as powerful tools for bridging the semantic gap between visual and textual modalities, enabling unprecedented open-vocabulary capabilities. Among these, CLIP~\cite{radford2021learning} stands out as a pivotal architecture due to its effective contrastive pre-training on a massive dataset of image-text pairs. CLIP comprises an image encoder and a text encoder, jointly trained to embed semantically related image-text pairs close to each other in a shared latent space.

For its image encoding, CLIP often leverages a Vision Transformer (ViT) architecture. In a ViT-based CLIP model, the image encoder processes an input image by dividing it into a sequence of fixed-size patches, which are then linearly embedded and combined with positional embeddings. A special learnable \texttt{[CLS]} token is prepended to this sequence, serving as a global representation for the entire image. This sequence of tokens is then fed through a series of Transformer blocks. Let $X=\left[ x_{cls}, x_1, ..., x_{h\times w} \right] ^{\mathsf{T}} \in \mathbb{R}^{(hw+1, d)}$ denote the input to the last Transformer block, where $h$ and $w$ represent the height and width of the feature map derived from the input image patches, and $d$ is the dimension of each token. The \texttt{[CLS]} token, $x_{cls}$, captures holistic image semantics, while $x_1, ..., x_{h\times w}$ represent local features corresponding to different image patches. The forward process of the last Transformer block can be formulated as follows:

\begin{equation}
\begin{aligned}
\begin{array}{c}
\boldsymbol{q}=\operatorname{Emb}_{q}(X), \boldsymbol{k}=\operatorname{Emb}_{k}(X), \boldsymbol{v}=\operatorname{Emb}_{v}(X), \\
\boldsymbol{y}=X+\operatorname{SA}\left(\boldsymbol{q}, \boldsymbol{k}, \boldsymbol{v}\right), \\
\boldsymbol{z}=\boldsymbol{y}+\operatorname{FFN}(\operatorname{LN}(\boldsymbol{y})),
\end{array}
\label{eq:attn}
\end{aligned}
\end{equation}
where $\boldsymbol{q}$, $\boldsymbol{k}$, and $\boldsymbol{v}$ denote the Query, Key, and Value matrices, respectively. $\operatorname{Emb}$ represents an embedding block typically composed of a Layer Normalization (LN) layer followed by a linear layer. $\operatorname{SA}$ signifies the standard multi-head self-attention module, mathematically expressed as $\operatorname{SA}(\boldsymbol{q}, \boldsymbol{k}, \boldsymbol{v}) = \text{softmax}(\frac{\boldsymbol{q} \cdot \boldsymbol{k}^\mathsf{T}}{\sqrt{d}}) \cdot \boldsymbol{v}$. The term $\sqrt{d}$ is a scaling factor to prevent large dot products. After the self-attention mechanism, the output $\boldsymbol{y}$ undergoes a feed-forward network (FFN) operation. Finally, a projection layer maps the refined token representations $\boldsymbol{z}$ to a multi-modal embedding space:

\begin{equation}
\begin{aligned}
\mathcal{O} = \operatorname{Proj}(\boldsymbol{z}),
\label{eq:proj}
\end{aligned}
\end{equation}
where $\mathcal{O} =\left[ o_{cls}, o_1, ..., o_{h\times w} \right] ^{\mathsf{T}} \in \mathbb{R}^{(hw+1, c)}$ denotes the final output of the image encoder. Here, $c$ is the token dimension after the projection layer, and typically $c < d$. In the original CLIP training, the global \texttt{[CLS]} token output, $o_{cls}$, is primarily used for image-level learning objectives (\textit{e.g.}, image-text matching). However, for downstream dense prediction tasks like OVSS, the local patch tokens, $\mathcal{O}[1:hw+1]$, are crucial for computing pixel-wise similarities with text embeddings.

% This distinction highlights a key challenge: the global \texttt{[CLS]} token's influence, while beneficial for image-level tasks, can sometimes introduce detrimental global biases into local patch representations, impacting fine-grained pixel-level accuracy.

\subsubsection{FeatUp: Model-Agnostic Feature Upsampling}
\label{sec:featup}

Dense prediction tasks, such as semantic segmentation, inherently require high-resolution outputs to accurately delineate object boundaries and fine-grained structures. However, modern VLMs like CLIP~\cite{radford2021learning} and DINO~\cite{caron2021emerging, oquab2023dinov2}, often operate on downsampled image features for computational efficiency. This necessitates an effective upsampling to bridge the resolution gap between the VLM's internal features and the desired high-resolution output. FeatUp \cite{fu2024featup} is a notable model-agnostic upsampler designed to address this requirement by training a universal upsampling module that can be applied to features from various frozen backbones.

FeatUp involves performing an upsampling operation on low-resolution features, denoted as $\mathcal{O}[1:hw+1]$ (from a frozen backbone), using a learnable upsampler $\sigma_{\uparrow}$. To ensure that the upsampled high-resolution features maintain consistency with the original low-resolution input, FeatUp also employs a learnable downsampler $\sigma_{\downarrow}$ to reconstruct the low-resolution features from the upsampled high-resolution output. The core training objective for FeatUp is to minimise the discrepancy between the original low-resolution features and their reconstructed version, defined by the following loss:

\begin{equation}
\begin{aligned}
\mathcal{L}_{rec}=\left\|\mathcal{O}[1:hw+1] - \sigma_{\downarrow}(\sigma_{\uparrow}(\mathcal{O}[1:hw+1])) \right\|_{2}^{2}.
\label{eq:featup_loss}
\end{aligned}
\end{equation}

FeatUp instantiates $\sigma_{\uparrow}$ using stacked parameterized JBU operators \cite{kopf2007joint}. JBU is a non-linear filter that estimates high-resolution pixels by weighting neighboring low-resolution pixels based on both spatial proximity and feature similarity in a guidance image. This allows JBU to preserve edges and fine details during upsampling, which is critical for maintaining perceptual quality. For weight generation, JBU considers two factors: $k_{range}$ (feature similarity-based kernel) and $k_{spatial}$ (spatial distance-based kernel), ensuring that more relevant and closer pixels contribute more significantly. It is important to note that FeatUp also incorporates a multi-view consistency constraint during its training, although for brevity, we omit its detailed formulation here. 

% FeatUp provides a strong foundation for model-agnostic upsampling, while our SimFeatUp, building upon these principles, is further refined to specifically address the unique requirements and challenges encountered in remote sensing OVSS, particularly concerning annotation-free applicability and content fidelity.

\subsection{SimFeatUp}
\label{sec:simfeatup}

Inspired by FeatUp \cite{fu2024featup}, which provides an excellent training paradigm for model-agnostic upsamplers, our SimFeatUp is specifically tailored to address the issue of annotation-free OVSS in remote sensing. Although FeatUp offers a general upsampling solution, it primarily considers settings where labels are available, and its direct application to annotation-free OVSS task, particularly within remote sensing contexts, yields sub-optimal performance due to specific limitations in content retention and feature utilization.

\subsubsection{Image Content Retention}
\label{sec:img_content_retention}

\begin{figure}[t]
  \centering
%   \fbox{\rule{0pt}{2in} \rule{0.9\linewidth}{0pt}}
   \includegraphics[width=0.85\linewidth]{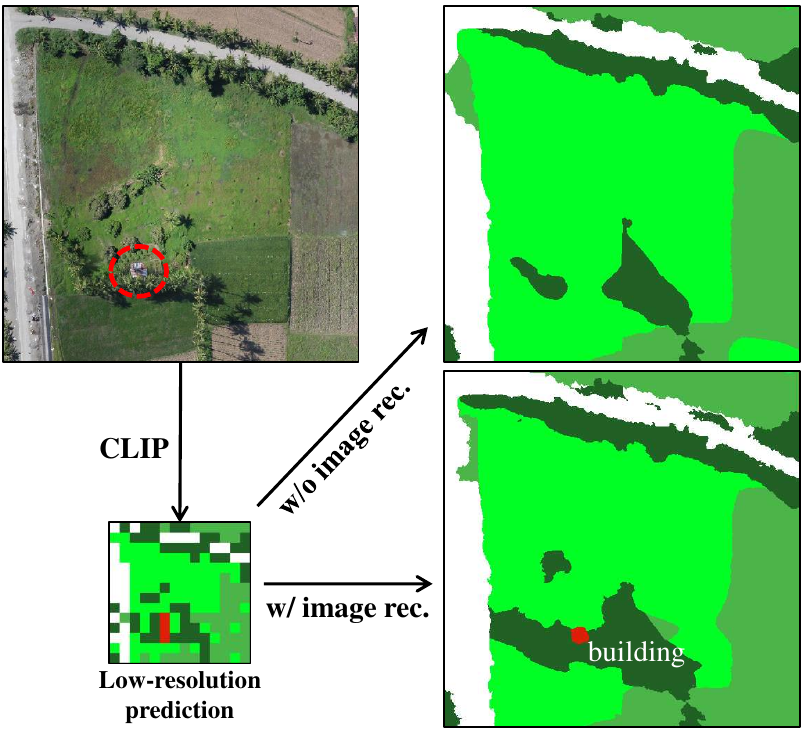}
   \caption{Comparison of with and without image reconstruction loss (\cref{eq:img_rec_loss}). The low-resolution prediction is obtained directly using the output of CLIP (without bilinear interpolation). Color: \textcolor[RGB]{222, 31, 7}{building}, \textcolor[RGB]{34, 97, 38}{tree}, \textcolor[RGB]{75, 181, 73}{cropland}, \textcolor[RGB]{0, 255, 36}{grass}.}
   \label{fig:fig_rec}
\vspace{-1em}
\end{figure}

As discussed in Section \ref{sec:featup}, FeatUp's primary objective is to minimize the reconstruction loss ($\mathcal{L}_{rec}$) between original low-resolution features and those reconstructed after an up-down-sampling cycle (\textit{i.e.}, $\sigma_{\downarrow}(\sigma_{\uparrow}(\mathcal{O}[1:hw+1]))$). However, since both the upsampler $\sigma_{\uparrow}$ and downsampler $\sigma_{\downarrow}$ are learnable, this constraint is relatively weak. The up-down-sampling process can operate as a ``black box'', potentially allowing the intermediate high-resolution features to become incomplete or inconsistent with the original image's semantic content. This can lead to a semantic drift, where fine details or small objects, initially present in the low-resolution prediction, might disappear or be distorted in the high-resolution output, as illustrated in Fig.~\ref{fig:fig_rec} (bottom left example, where a small building vanishes).

To overcome this crucial issue and ensure that the generated high-resolution features are semantically complete and visually consistent with the input image, we introduce an additional content retention network, guided by a reconstruction loss ($\mathcal{L}_{img}$). This loss directly constrains the upsampled high-resolution features by forcing them to reconstruct the original input image. The $\mathcal{L}_{img}$ is formulated as:

\begin{equation}
\begin{aligned}
\mathcal{L}_{img}=\left\| I - \operatorname{CRN}(\sigma_{\uparrow}(\mathcal{O}[1:hw+1])) \right\|_{2}^{2},
\label{eq:img_rec_loss}
\end{aligned}
\end{equation}
where $I$ denotes the original input image, and $\operatorname{CRN}$ is the content retention network. The $\operatorname{CRN}$ receives the upsampled high-resolution features ($\sigma_{\uparrow}(\mathcal{O}[1:hw+1]))$ as input and aims to reconstruct the original image $I$. Specifically, $\operatorname{CRN}$ is designed with two convolutional layers, followed by normalization layers and a \textit{Tanh} activation function. The \textit{Tanh} activation constrains the output pixel values within the range [-1, 1], similar to its usage in VAEs~\cite{kingma2013auto} for image generation. This explicit constraint ensures that the spatial details and semantic content, critical for remote sensing tasks, are faithfully preserved throughout the upsampling process. Finally, the total loss function for training SimFeatUp combines both the feature reconstruction loss and the image reconstruction loss with a balancing weight $\gamma$:

\begin{equation}
\begin{aligned}
\mathcal{L} = \mathcal{L}_{rec} + \gamma \mathcal{L}_{img}.
\label{eq:loss}
\end{aligned}
\end{equation}
% This combined objective not only ensures feature-level consistency but also enforces visual fidelity, leading to more accurate and reliable high-resolution feature maps for downstream OVSS.

\subsubsection{Optimal Feature Selection for Upsampling}
\label{sec:which_feature_to_upsample}

FeatUp \cite{fu2024featup} typically uses the final output of CLIP's image encoder, $\mathcal{O}[1:hw+1]$ (as defined in \cref{eq:proj}), as the input to its upsampler. This choice can be effective in training-based scenarios, such as linear probing \cite{alain2016understanding}. However, in annotation-free OVSS, as described in \cref{sec:related_work_ovss}, vanilla self-attention leads to inferior performance. Therefore, the current OVSS method modulates the last self attention to self-self attention to improve pixel-level localization capability, and this law also works in remote sensing images. Under this premise, the $\operatorname{SA}$ in \cref{eq:attn} would be replaced by other modules, and direct upsampling of $\mathcal{O}[1:hw+1]$ would lead to the mismatch between SimFeatUp's training and its specific application within the OVSS inference pipeline.

Motivated by this, we propose to leverage CLIP features from an earlier stage for upsampling. Specifically, we choose the input to the last Transformer block of CLIP's image encoder, $X[1:hw+1]$ (as shown in \cref{eq:attn}).  Furthermore, the high dimension of tokens in $X$ leads to a high-cost upsampler. Therefore, we retain CLIP's original projection layer. Ultimately, the features $\mathcal{O}^{\prime}$ that serve as input to our SimFeatUp upsampler are formulated as:

\begin{equation}
\begin{aligned}
\mathcal{O}^{\prime} = \operatorname{Proj}(X[1:hw+1]).
\label{eq:which_fea}
\end{aligned}
\end{equation}

\subsubsection{Larger Upsampling Kernel for Remote Sensing Objects}
\label{sec:larger_upsampling_kernel}

Our SimFeatUp adopts the parameterized JBU operator, similar to FeatUp \cite{fu2024featup}. As elaborated in Section \ref{sec:featup}, JBU computes upsampling kernels based on both spatial distance ($k_{spatial}$) and feature similarity ($k_{range}$) within a local window in a guidance feature. The formulations for these kernels are as follows:

\begin{equation}
\begin{aligned}
k_{spatial}(p, q)=\exp \left(\frac{-\|p-q\|_{2}^{2}}{2 \tau_{spatial}^{2}}\right),
\label{eq:k_spatial}
\end{aligned}
\vspace{-4mm}
\end{equation}

\begin{equation}
\begin{aligned}
&k_{range}(p, q) = \\
&\operatorname{softmax}_{(a, b) \in \Omega}\left(\frac{1}{\tau_{range}^{2}} \textit{MLP}(G[i, j]) \cdot \textit{MLP}(G[a, b])\right),
\label{eq:k_range}
\end{aligned}
\end{equation}
where $(p, q)$ denotes the position within the kernel, $\Omega$ represents a local window centered at $(i, j)$ in the guidance feature $G$ (extracted from the high-resolution RGB image), and $\tau_{spatial}$ and $\tau_{range}$ are learnable factors controlling the influence of spatial distance and feature similarity.

A key consideration in remote sensing is the vast and often logarithmic scale variation of targets \cite{rolf2024mission}. Objects can range from small trees and gardens to expansive forests and rangelands. To effectively capture these diverse scales and ensure a sufficiently wide receptive field for large objects, we expanded the default JBU window size from $7 \times 7$ (as in FeatUp) to $11 \times 11$. Although a larger receptive field might intuitively introduce more irrelevant context, the adaptive nature of $k_{spatial}$ mitigates this concern. $k_{spatial}$ assigns lower weights to more distant points, meaning that distant yet semantically irrelevant pixels contribute minimally to the upsampled output. This design makes the larger kernel size reasonable and beneficial, allowing SimFeatUp to capture multi-scale objects while maintaining local fidelity effectively.

\subsubsection{Structural Simplification}
\label{sec:structural_simplification}

In its original design, FeatUp \cite{fu2024featup} often employs stacked parameterized JBU modules (\textit{e.g.}, 4 modules for $16\times$ upsampling), where each module operates independently with its own set of parameters. Although effective, this modularity can introduce complexities in training and make the behaviour of each independent JBU module less determinable, even with the $\operatorname{CRN}$ ensuring content integrity. Therefore, we simplify the structural components of SimFeatUp. Instead of a stacked "JBU\_Stack" approach, we consolidate it into a single parameterized "JBU\_One" module. This means only one learnable JBU module is used for upsampling. If a $16\times$ upsampling ratio is required, this single "JBU\_One" module is executed 4 times sequentially. This simplification offers several advantages:
\begin{itemize}
    \item Reduced Trainable Parameters: Consolidating multiple JBU modules into one significantly reduces the total number of trainable parameters in the upsampler, making SimFeatUp more lightweight.
    \item Improved Determinism: A single, shared JBU module across all upsampling steps leads to more predictable and stable behaviour during inference.
    \item Flexible Upsampling Ratios: The "JBU\_One" design inherently supports upsampling by arbitrary multiples (\textit{e.g.}, $2\times, 4\times, \dots, 16\times$) by simply repeating its execution the required number of times, enhancing its versatility.
\end{itemize}
% This structural simplification not only makes SimFeatUp more efficient and robust but also aligns perfectly with our goal of a training-agnostic upsampler that can be seamlessly integrated into various remote sensing OVSS pipelines, including those involving SAR imagery.

\subsection{Global Bias Alleviation}
\label{sec:global_bias_alleviation}

Pre-trained VLMs, while powerful in aligning visual and textual modalities, often exhibit an inherent ``global bias'' that can be detrimental to fine-grained dense prediction tasks like semantic segmentation. As detailed in Section \ref{sec:clip}, during CLIP's training phase, the \texttt{[CLS]} token is optimized to align with the corresponding text embedding in the multi-modal space via contrastive learning. However, a significant discrepancy arises during OVSS inference: the \texttt{[CLS]} token is generally discarded, and only local patch tokens ($\mathcal{O}[1:hw+1]$) are utilized for similarity computation with prompt vocabularies. This divergence between training (global focus) and inference (local focus) creates a ``gap'' that can lead to suboptimal performance in dense prediction.

Previous research \cite{mukhoti2023open, ranasinghe2023perceptual, wang2023sclip} has empirically demonstrated that each local visual token in CLIP tends to focus on a wide range of positions, and their attention maps often exhibit similar, non-discriminative patterns across different regions. This phenomenon strongly suggests that global attributes, captured by the \texttt{[CLS]} token, are ``attached'' or ``leaked'' into the local patch tokens. Although this property is generally not a concern for image-level classification tasks, where overall image content is paramount, it significantly impairs performance in dense prediction for tasks requiring precise pixel-level understanding. The global context might inadvertently activate irrelevant local regions, leading to inaccurate segmentation, especially for objects that are visually distinct from the dominant global theme of the image.

\begin{figure}[t]
  \centering
   \includegraphics[width=0.95\linewidth]{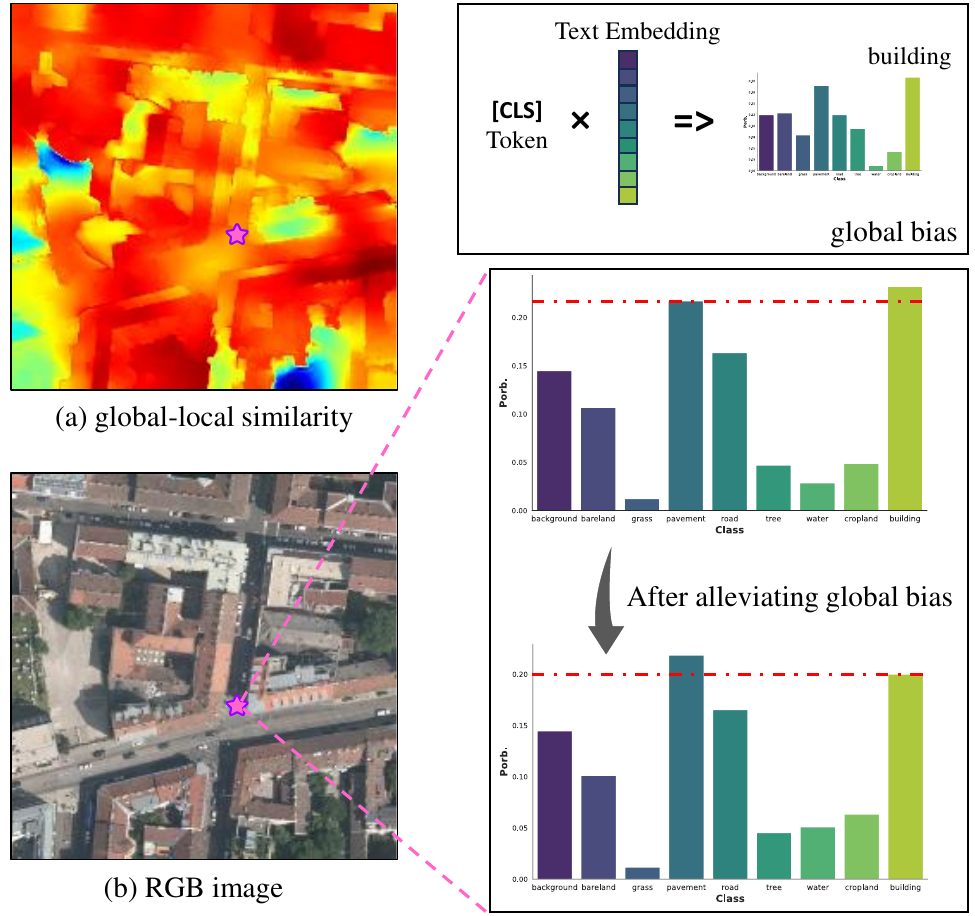}
   \caption{Comparison of before and after alleviating the global bias. (a) is the similarity map of patch tokens and \texttt{[CLS]} tokens, showing that some "non-building" regions also present high response. (b) is the original RGB image. Note that the right-hand histograms stretch the raw values for better presentation.}
   \label{fig:fig_global_bias}
   \vspace{-1em}
\end{figure}

The visualization in Fig.~\ref{fig:fig_global_bias} demonstrates the above elaboration. We extract the \texttt{[CLS]} token from CLIP for the RGB image in Fig.~\ref{fig:fig_global_bias} (b) and compute its cosine similarity with various candidate text embeddings. The image is correctly recognized as the ``building'', which is reasonable given that the buildings occupy a significant portion of the image. Subsequently, we calculate the similarity map between the \texttt{[CLS]} token and each local patch token, as presented in Fig. \ref{fig:fig_global_bias} (a). The visualization clearly shows that highly responsive regions are not limited to actual building areas; rather, some non-building regions, such as roads and pavements, also exhibit high activation. This strong activation in semantically unrelated areas is a direct indicator that the global bias contaminates the local patch tokens, leading to a loss of fine-grained local discriminability.

Motivated by this observation, we propose a simple yet highly effective solution, Global Bias Alleviation. This strategy involves a straightforward subtraction operation to explicitly reduce the influence of global bias from the patch tokens. The corrected patch tokens, $\hat{\mathcal{O}}$, are formulated as:

\begin{equation}
\begin{aligned}
\hat{\mathcal{O}} = \mathcal{O}[1:hw+1] - \lambda\mathcal{O}[0],
\label{eq:global}
\end{aligned}
\end{equation}
where $\mathcal{O}[1:hw+1]$ represents the raw patch tokens from CLIP's output, $\mathcal{O}[0]$ is the \texttt{[CLS]} token which is spatially replicated to match the dimensions of the patch tokens, and $\lambda$ is an intensity factor. This scaling factor $\lambda$ controls the degree to which the global bias is suppressed. By performing this subtraction, we explicitly encourage the model to rely more on the intrinsic local visual cues and fine-grained textures, rather than being overly influenced by the image's overall semantic content. This significantly improves local semantic fidelity and increases the accuracy of pixel-level predictions. Global Bias Alleviation operates as a post-processing step without requiring additional training or fine-tuning of the VLM, making it a highly practical and efficient component of our SegEarth-OV framework.

\section{AlignEarth: Cross-modal Remote Sensing Knowledge Distillation}
\label{sec:alignearth}

The ambition of remote sensing OVSS extends beyond optical image to encompass diverse remote sensing modalities, \textit{e.g.}, SAR, infrared images, \textit{etc}. Adapting OVSS to these heterogeneous sensor types, however, poses distinct challenges due to their different imaging mechanisms and the scarcity of modality-specific pre-trained VLMs. To address this, we propose AlignEarth, a cross-modal remote sensing knowledge distillation strategy. AlignEarth is designed to effectively transfer rich semantic knowledge from a well-established source modality (\textit{e.g.}, optical VLMs) to a target sensor modality (\textit{e.g.}, SAR) where pre-trained VLMs or extensive annotations are scarce. Although applicable to various sensor types, this paper primarily focuses on its practical application to SAR images, leveraging paired optical-SAR images to distil knowledge from optical VLMs into a dedicated SAR image encoder.

\subsection{SAR Images in Remote Sensing}
\label{sec:sar_characteristics_challenges}

\begin{figure}[t]
  \centering
   \includegraphics[width=1.0\linewidth]{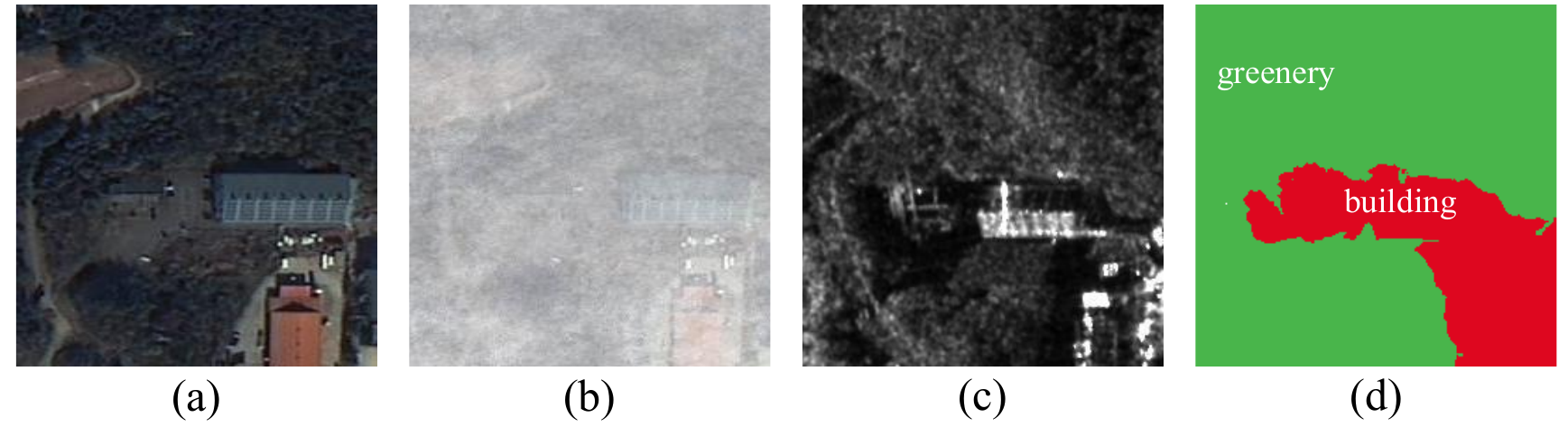}
   \caption{Different remote sensing image modalities. (a) optical image w/o cloud. (b) optical image w/ cloud. (c) SAR image. (d) our prediction (SegEarth-OV + AlignEarth) using only the SAR image.}
   \label{fig:fig_sar_img}
   % \vspace{-1em}
\end{figure}

SAR image constitutes a critical component of comprehensive Earth observation, offering unique advantages over optical remote sensing, particularly its all-weather, all-day operability and penetration capabilities through clouds, smoke, and certain vegetation layers. These attributes make SAR indispensable for applications like disaster monitoring, land cover mapping under adverse weather conditions, and precise terrain analysis. However, SAR images are fundamentally different from optical images, both in their acquisition mechanism and visual appearance, leading to distinct challenges for OVSS, as shown in Fig.~\ref{fig:fig_sar_img}.

One of the primary distinctions lies in SAR's active sensing and backscattering principle. Unlike passive optical sensors that record reflected sunlight, SAR systems actively emit microwave pulses and record the backscattered signals. The intensity of SAR pixels represents the radar cross-section of the illuminated targets, which depends on geometric properties (shape, orientation), material dielectric properties (conductivity, moisture content), and surface roughness. This results in textural patterns and intensity variations that are highly distinct from optical images. Furthermore, a ubiquitous characteristic of SAR images is speckle noise, a granular appearance caused by the constructive and destructive interference of scattered waves within a resolution cell. Speckle is a multiplicative noise that significantly degrades image quality, obscures fine details, and complicates feature extraction, posing a considerable challenge for pixel-level analysis. Beyond these, SAR images inherently lack color information, relying solely on intensity and polarization to convey information. The textures are derived from radar backscatter properties rather than visual appearance, which often makes them less intuitive for human interpretation and, consequently, for VLMs trained predominantly on natural, color-rich optical images.

Given these unique characteristics, directly applying VLMs pre-trained on optical images (like CLIP) to SAR images for OVSS proves largely ineffective, as shown in Table~\ref{table_sar_vs_rsclip}. The modality gap means that the visual representations learned by optical VLMs are not inherently compatible with SAR data. Moreover, there is currently no publicly available large-scale pre-trained VLM analogous to CLIP specifically designed for SAR images. This absence stems from the immense difficulty and prohibitive cost of collecting and annotating massive SAR image-text pairs, which would be necessary for such pre-training. Generating high-quality, semantically rich textual descriptions for SAR images is more challenging than for optical images. The nature of SAR backscatter values makes it difficult for human annotators or Multi-modal Large Language Models (MLLMs) to consistently and precisely describe objects and scenes in a textual form that aligns with the deep semantic understanding required by VLMs. These multifaceted challenges constitute a significant barrier to effectively extending OVSS capabilities to SAR images. 

% AlignEarth can address this critical gap by leveraging existing knowledge from optical VLMs to effectively train an SAR image encoder without requiring large volumes of new, expensive SAR-specific text annotations.

\subsection{Training Framework}
\label{sec:alignearth_training_framework}

% To bridge the significant modality gap and enable open-vocabulary semantic segmentation for SAR images without requiring SAR-specific textual annotations, we propose AlignEarth, a simple cross-modal knowledge distillation framework. AlignEarth efficiently transfers rich semantic knowledge from a powerful, pre-trained optical VLM (specifically, CLIP's image and text encoders) to a dedicated SAR image encoder using paired optical-SAR images. This framework comprises a frozen optical CLIP image encoder as the teacher and a trainable SAR image encoder as the student, guided by a multi-faceted loss function.

To bridge the modality gap and transfer rich semantic knowledge from pre-trained optical VLMs to the SAR domain without requiring SAR-specific text annotations, we propose AlignEarth, which leverages paired optical-SAR images to efficiently train a dedicated SAR image encoder, enabling it to produce representations semantically aligned with the optical CLIP embedding space.

The core of AlignEarth's training framework involves two main components: a frozen CLIP image encoder acting as the teacher, denoted as $\mathcal{E}_{opt}$, and a trainable SAR image encoder serving as the student, denoted as $\mathcal{E}_{sar}$. During training, we feed paired optical-SAR images $(I_{opt}, I_{sar})$ into their respective encoders. The optical image $I_{opt}$ is processed by the frozen $\mathcal{E}_{opt}$, which has already learned powerful visual-textual alignments from vast natural image datasets. The SAR image $I_{sar}$ is processed by the trainable $\mathcal{E}_{sar}$. Our objective is to distill the knowledge from $\mathcal{E}_{opt}$ to $\mathcal{E}_{sar}$, allowing $\mathcal{E}_{sar}$ to generate SAR features that are semantically consistent with CLIP's optical representations, as shown in Fig.~\ref{fig:method} (c).

The training objective for AlignEarth is formulated through a hybrid loss, designed to capture both global semantic alignment and fine-grained local feature consistency. The total loss consists of global contrastive loss, global distillation loss, and local distillation loss.

\subsubsection{Global Contrastive Loss}

Drawing inspiration from CLIP's training paradigm, we employ a contrastive learning objective on the global tokens. This loss encourages the SAR encoder to learn discriminative representations that align its global SAR image semantics with the corresponding optical image semantics. For a given pair $(I_{opt}, I_{sar})$, we obtain the final projected global tokens (\textit{i.e.}, \texttt{[CLS]} token, specifically ``$\mathcal{O}[0]$'' as defined in Section \ref{sec:clip}) from both encoders: $\mathcal{O}_{opt\_cls} = \mathcal{E}_{opt}(I_{opt})[0]$ and $\mathcal{O}_{sar\_cls} = \mathcal{E}_{sar}(I_{sar})[0]$. In a mini-batch of $N$ paired images, we treat $(\mathcal{O}_{opt\_cls}^{(i)}, \mathcal{O}_{sar\_cls}^{(i)})$ from the same pair as a positive sample. All other $N-1$ SAR class tokens from different pairs within the batch serve as negative samples for $\mathcal{O}_{opt\_cls}$, and vice versa. We optimize a symmetric cross-entropy loss over the similarity scores of these pairs:

{\footnotesize
\begin{equation}
\label{eq:cls_contrast_loss}
\begin{split}
\mathcal{L}_{cls\_contrast} = -\frac{1}{N} \sum_{i=1}^{N} & \left[ \log \frac{\exp(\text{sim}(\mathcal{O}_{opt\_cls}^{(i)}, \mathcal{O}_{sar\_cls}^{(i)}) / \tau)}{\sum_{j=1}^{N} \exp(\text{sim}(\mathcal{O}_{opt\_cls}^{(i)}, \mathcal{O}_{sar\_cls}^{(j)}) / \tau)} \right. \\
& \quad \left. + \log \frac{\exp(\text{sim}(\mathcal{O}_{sar\_cls}^{(i)}, \mathcal{O}_{opt\_cls}^{(i)}) / \tau)}{\sum_{j=1}^{N} \exp(\text{sim}(\mathcal{O}_{sar\_cls}^{(i)}, \mathcal{O}_{opt\_cls}^{(j)}) / \tau)} \right]
\end{split}
\end{equation}
}where $\text{sim}(\cdot, \cdot)$ denotes the cosine similarity, and $\tau$ is a learnable temperature parameter.
This loss encourages $\mathcal{E}_{sar}$ to learn discriminative representations, pulling its global SAR image semantics closer to the corresponding optical image semantics while pushing it away from unrelated ones.

% This loss enables $\mathcal{E}_{sar}$ to learn a global understanding of SAR scenes that is discriminative and semantically consistent with the global representations produced by $\mathcal{E}_{opt}$.

\subsubsection{Global Distillation Loss}

Beyond discriminative alignment, we also explicitly enforce the SAR features to achieve semantic consistency with their optical counterparts in the shared embedding space. This is realized through a direct distillation loss on the class tokens using cosine similarity. Unlike the contrastive loss, which focuses on distinguishing positive from negative pairs, this distillation loss directly pulls the features of corresponding positive pairs closer in terms of their vector orientation, aiming for maximal cosine similarity. This ensures that the global semantic information extracted from SAR images closely mirrors that from optical images, effectively aligning their overall semantic content in the embedding space.

\begin{equation}
\mathcal{L}_{cls\_distill} = 1 - \text{sim}(\mathcal{O}_{opt\_cls}, \mathcal{O}_{sar\_cls})
\label{eq:cls_distill_loss}
\end{equation}
This loss acts as a direct regularization, ensuring that the global semantic information extracted from SAR images closely mirrors that from optical images. Minimizing this loss helps in achieving a more direct and precise alignment of the SAR feature space with the optical CLIP feature space.

\begin{table*}
  \caption{The prompt class name of the evaluation datasets. \{\} indicates multiple prompt vocabularies for one class.}
  \label{table_class_name}
  \centering
  \scalebox{0.95}{
  \begin{tabular}{@{}p{3.7cm}>{\centering\arraybackslash}m{15cm}@{}}
    \toprule[1pt]
    Dataset & Class Name \\
    \midrule[1pt]
    OpenEarthMap & background, \{bareland, barren\}, 
grass, pavement, road, \{tree, forest\}, \{water, river\}, cropland, \{building, roof, house\} \\
    LoveDA & background, \{building, roof, house\}, road, water, barren, forest, agricultural \\
    iSAID &  background, ship, store tank, baseball diamond, tennis court, basketball court, ground track field, bridge, large vehicle, small vehicle, helicopter, swimming pool, roundabout, soccer ball field, plane, harbor \\
    Potsdam, Vaihingen &  road, building, grass, tree, car, \{clutter, background\} \\
    UAVid & background, building, road, car, tree, vegetation, human \\
    UDD5 & vegetation, building, road, vehicle, background \\
    VDD & background, facade, road, vegetation, vehicle, roof, water \\
    \midrule[1pt]
    WHU$^{Aerial}$, WHU$^{Sat.\mathrm{II}}$, \\ Inria, xBD & background, building \\
    CHN6-CUG, DeepGlobe, \\ Massachusetts, SpaceNet & background, road \\
    WBS-SI & background, water \\
    \midrule[1pt]
    PIE-SAR & background, city, road, \{water,river\}, forest, cropland\\
    YESeg-SAR & \{bareground,barren\}, \{grass,farmland\}, dense tree cover, city, water, roadway\\
    FUSAR-Map & others, water, road, building, grass\\
    DDHR-Korea, DDHR-SD, \\ DDHR-XA & building, road, greenery, water, \{farmland,grass\} \\ 
    WHU-SAR & farmland, city, village, water, forest, road, others\\
    OpenEarthMap-SAR & background, \{bareland, barren\}, 
grass, pavement, road, \{tree, forest\}, \{water, river\}, cropland, \{building, roof, house\} \\
    \bottomrule[1pt]
  \end{tabular}}
\end{table*}

\subsubsection{Local Distillation Loss}

Although global semantic alignment is crucial, pixel-level semantic segmentation demands fine-grained local feature understanding. Both $\mathcal{E}_{opt}$ and $\mathcal{E}_{sar}$ produce final projected feature maps containing local patch tokens that correspond to spatial regions within the input image. Let $\mathcal{O}_{opt\_local} = \mathcal{E}_{opt}(I_{opt})[1:hw+1]$ and $\mathcal{O}_{sar\_local} = \mathcal{E}_{sar}(I_{sar})[1:hw+1]$ be the local feature maps. However, a direct token-wise distillation on these local feature maps is problematic. Paired optical-SAR datasets, despite best efforts, often suffer from: (1) imperfect geometric co-registration, where a pixel in the optical image does not perfectly align with its semantic equivalent in the SAR image, and (2) modality-specific distortions (\textit{e.g.}, radar shadows, layover) that create semantic inconsistencies at the local level. These issues make a direct one-to-one token distillation noisy and unreliable.

% Directly applying pixel-wise or token-wise distillation (\textit{e.g.}, L2 or cosine similarity loss) on these raw local feature maps is problematic due to the inherent challenges in optical-SAR image pairing. Despite efforts in data collection, paired optical-SAR datasets often suffer from ``imperfect geometric co-registration'' and ``modality-specific distortions'' (\textit{e.g.}, radar shadow, layover), meaning that a pixel in the optical image may not perfectly correspond to the semantically equivalent pixel in the SAR image.

To address this, we propose a robust region-level distillation strategy that operates on a coarser granularity. We divide both $\mathcal{O}_{opt\_local}$ and $\mathcal{O}_{sar\_local}$ into $K \times K$ non-overlapping regions. For each region at row $i$ and column $j$, denoted as $P_{opt}^{(i,j)}$ for $\mathcal{O}_{opt\_local}$ and $P_{sar}^{(i,j)}$ for $\mathcal{O}_{sar\_local}$, we compute its mean feature vector: $\bar{f}_{opt\_patch}^{(i,j)} = \text{mean}(P_{opt}^{(i,j)})$ and $\bar{f}_{sar\_patch}^{(i,j)} = \text{mean}(P_{sar}^{(i,j)})$. This averaging operation effectively mitigates the impact of minor misalignments and speckle noise, providing a more robust local descriptor for distillation. The local distillation loss is then computed as the cosine distance between corresponding local descriptors:

\begin{equation}
\mathcal{L}_{local\_distill} = \frac{1}{K^2} \sum_{i=1}^{K} \sum_{j=1}^{K} (1 - \text{sim}(\bar{f}_{opt\_patch}^{(i,j)}, \bar{f}_{sar\_patch}^{(i,j)}))
\label{eq:local_distill_loss}
\end{equation}
This loss ensures that the fine-grained semantic information from local regions in SAR images is aligned with that of optical images.
% , but in a way that is robust to strict pixel-level correspondences.

Once the SAR image encoder $\mathcal{E}_{sar}$ is trained with AlignEarth, it is frozen and serves as the CLIP's image encoder for SAR images within the SegEarth-OV framework. The outputs of $\mathcal{E}_{sar}$ (both \texttt{[CLS]} and patch tokens) are then \textbf{directly fed into our existing SimFeatUp and Global Bias Alleviation modules}. Crucially, SimFeatUp can directly process the features from $\mathcal{E}_{sar}$ without requiring any re-training. Similarly, Global Bias Alleviation operates effectively on these SAR features, mitigating global biases and enhancing local semantic fidelity. This seamless integration ensures that SegEarth-OV has the potential to universally perform OVSS across diverse remote sensing modalities.

\section{Experiments}
\label{sec:exps}

This section presents a comprehensive evaluation of our proposed SegEarth-OV framework, encompassing both optical and SAR remote sensing images. We first detail the experimental setup, including datasets, evaluation metrics, and implementation details. Subsequently, we present quantitative and qualitative results, comparing SegEarth-OV against SOTA annotation-free OVSS methods on optical images, and demonstrating its capabilities on SAR images. Finally, we conduct extensive ablation studies to analyze the contribution of each proposed component, including SimFeatUp, Global Bias Alleviation, and AlignEarth.

\begin{table*}
  \caption{Open-vocabulary semantic segmentation quantitative comparison on remote sensing datasets. Evaluation metric: mIoU. \textcolor{tabred}{\textbf{Best}} and \textcolor{tabblue}{\textbf{second best}} performances are highlighted. ``Oracle'' is achieved by a fully supervised SegFormer~\cite{xie2021segformer} model using full training data, representing the upper bound.}
  \label{table_main}
  \centering
  \scalebox{1.0}{
  \begin{tabular}{@{}llcccccccc|c@{}}
    \toprule[1pt]
    Methods & & OpenEarthMap & LoveDA & iSAID & Potsdam & Vaihingen & UAVid$^{img}$ & UDD5 & VDD & Average \\
    \midrule[1pt]
    CLIP \cite{radford2021learning} & {\tiny ICML'21} & 12.0 & 12.4 & 7.5 & 15.6 & 10.8 & 10.9 & 9.5 & 14.2 & 11.4 \\
    MaskCLIP \cite{zhou2022extract} & {\tiny ECCV'22} & 25.1 & 27.8 & 14.5 & 33.9 & 29.9 & 28.6 & 32.4 & 32.9 & 27.2 \\
    % CLIPSurgery&  &  &  &  &\\
    SCLIP \cite{wang2023sclip} & {\tiny arXiv'23} & 29.3 & 30.4 & 16.1 & 39.6 & 35.9 & 31.4 & 38.7 & 37.9 & 31.1 \\
    GEM \cite{bousselham2024grounding} & {\tiny CVPR'24} & \textcolor{tabblue}{\textbf{33.9}} & 31.6 & 17.7 & 39.1 & \textcolor{tabblue}{\textbf{36.4}} & 33.4 & 41.2 & \textcolor{tabblue}{\textbf{39.5}} & 32.3 \\
    ClearCLIP \cite{lan2024clearclip} & {\tiny ECCV'24} & 31.0 & \textcolor{tabblue}{\textbf{32.4}} & \textcolor{tabblue}{\textbf{18.2}} & \textcolor{tabblue}{\textbf{42.0}} & 36.2 & \textcolor{tabblue}{\textbf{36.2}} & \textcolor{tabblue}{\textbf{41.8}} & 39.3 & \textcolor{tabblue}{\textbf{33.4}} \\
    SegEarth-OV & {\tiny Ours} & \textcolor{tabred}{\textbf{40.3}} & \textcolor{tabred}{\textbf{36.9}} & \textcolor{tabred}{\textbf{21.7}} & \textcolor{tabred}{\textbf{48.5}} & \textcolor{tabred}{\textbf{40.0}} & \textcolor{tabred}{\textbf{42.5}} & \textcolor{tabred}{\textbf{50.6}} & \textcolor{tabred}{\textbf{45.3}} & \textcolor{tabred}{\textbf{39.2}} \\
    \midrule[1pt]
    \textcolor{tablegray}{Oracle} & & \textcolor{tablegray}{64.4} & \textcolor{tablegray}{50.0} & \textcolor{tablegray}{36.2} & \textcolor{tablegray}{74.3} & \textcolor{tablegray}{61.2} & \textcolor{tablegray}{59.7} & \textcolor{tablegray}{56.5} & \textcolor{tablegray}{62.9} & \textcolor{tablegray}{58.2} \\
    \bottomrule[1pt]
  \end{tabular}}
  \vspace{-1em}
\end{table*}

\begin{table*}
  \caption{Open-vocabulary building/road/flood extraction quantitative comparison on remote sensing datasets. Evaluation metric: IoU of the foreground class, \textit{i.e.} building, road or flood. \textcolor{tabred}{\textbf{Best}} and \textcolor{tabblue}{\textbf{second best}} performances are highlighted.}
  \label{table_main_br}
  \centering
  \scalebox{0.95}{
  \begin{tabular}{@{}lcccc|cccc|c@{}}
    \toprule[1pt]
    \multirow{2}{*}{Method} & \multicolumn{4}{c|}{\color{red!50!black}Building Extraction} & \multicolumn{4}{c|}{\color{yellow!50!black}Road Extraction} & \multicolumn{1}{c}{\color{blue!50!black}Flood Detection}\\
    & WHU$^{Aerial}$ & WHU$^{Sat.\mathrm{II}}$  & Inria & xBD$^{pre}$ & CHN6-CUG & DeepGlobe & Massachusetts & SpaceNet & WBS-SI \\
    \midrule[1pt]
    \textbf{\textit{448 $\times$ 448:}} & & & & & & & & & \\
    CLIP \cite{radford2021learning} & 17.7 & 3.5 & 19.6 & 16.0 & 7.7 & 3.9 & 4.9 & 7.1 & 18.6 \\
    MaskCLIP \cite{zhou2022extract} & 29.8 & 14.0 & 33.4 & 29.2 & 28.1 & 13.2 & 10.6 & 20.8 & 39.8 \\
    SCLIP \cite{wang2023sclip} & 33.4 & 21.0 & 34.9 & 25.9 & 21.1 & 7.0 & 7.4 & 14.9 & 32.1 \\
    GEM \cite{bousselham2024grounding} & 24.4 & 13.6 & 28.5 & 20.8 & 13.4 & 4.7 & 5.1 & 11.9 & 39.5 \\
    ClearCLIP \cite{lan2024clearclip} & 36.6 & \textcolor{tabblue}{\textbf{20.8}} & 39.0 & 30.1 & 25.5 & 5.7 & 6.4 & 16.3 & 44.9\\
    SegEarth-OV & \textcolor{tabblue}{\textbf{49.2}} & \textcolor{tabred}{\textbf{28.4}} & \textcolor{tabblue}{\textbf{44.6}} & \textcolor{tabblue}{\textbf{37.0}} & \textcolor{tabred}{\textbf{35.4}} & \textcolor{tabblue}{\textbf{17.8}} & \textcolor{tabblue}{\textbf{11.5}} & \textcolor{tabblue}{\textbf{23.8}} & \textcolor{tabred}{\textbf{60.2}} \\
    \midrule[1pt]
    \textbf{\textit{896 $\times$ 896:}} & & & & & & & & & \\
    SegEarth-OV & \textcolor{tabred}{\textbf{49.9}} & - & \textcolor{tabred}{\textbf{48.9}} & \textcolor{tabred}{\textbf{43.1}} & 32.8 & \textcolor{tabred}{\textbf{20.1}} & \textcolor{tabred}{\textbf{17.2}} & \textcolor{tabred}{\textbf{29.1}} & \textcolor{tabblue}{\textbf{57.9}} \\
    \bottomrule[1pt]
  \end{tabular}}
  \vspace{-1em}
\end{table*}

\subsection{Dataset}

% In remote sensing application contexts, not only multi-class semantic segmentation but also extraction of certain land cover types (\textit{e.g.}, buildings, roads, water bodies) is required, \textit{e.g.}, Google's Open Buildings project\footnote{https://sites.research.google/gr/open-buildings}. Therefore, we select 17 typical datasets covering common semantic segmentation, building extraction, road extraction, and water body segmentation (flood detection) tasks.

Our experiments cover a wide range of remote sensing tasks and modalities, including optical image multi-class semantic segmentation, single-class extraction (buildings, roads, water bodies), and SAR image semantic segmentation.

\noindent
\textbf{Optical Remote Sensing Datasets.} We evaluate SegEarth-OV on 17 typical optical remote sensing datasets, broadly categorised into semantic segmentation, building extraction, road extraction, and flood detection tasks. These datasets capture diverse scene types, resolutions, and object scales, ensuring a comprehensive assessment of our method's performance on optical data.

\begin{itemize}
    \item \textbf{Semantic Segmentation:} We utilize 8 datasets: OpenEarthMap \cite{xia2023openearthmap}, LoveDA \cite{loveda}, iSAID \cite{waqas2019isaid}, Potsdam, Vaihingen\footnote{https://www.isprs.org/education/benchmarks/UrbanSemLab}, UAVid \cite{LYU2020108}, UDD5 \cite{chen2018large}, and VDD \cite{cai2023vdd}. The first five primarily contain satellite and aerial images, while the latter three contain UAV images.
    
    \item \textbf{Single-Class Extraction:} We select 4 building extraction datasets (\textit{i.e.}, WHU$^{Aerial}$ \cite{ji2018fully}, WHU$^{Sat.\mathrm{II}}$ \cite{ji2018fully}, Inria \cite{maggiori2017can}, and xBD \cite{gupta2019xbddatasetassessingbuilding}), 4 road extraction datasets (\textit{i.e.}, CHN6-CUG \cite{zhu2021global}, DeepGlobe \cite{demir2018deepglobe}, Massachusetts \cite{MnihThesis}, and SpaceNet \cite{van2018spacenet}), and 1 flood detection dataset (\textit{i.e.}, WBS-SI\footnote{https://www.kaggle.com/datasets/shirshmall/water-body-segmentation-in-satellite-images}) for the evaluation of single-class extraction. These datasets contain 1 foreground class (building, road or flood) and 1 background class.

\end{itemize}

\noindent
\textbf{SAR Remote Sensing Datasets.} To validate AlignEarth and extend SegEarth-OV's capabilities to SAR images, we conduct experiments on 8 SAR segmentation datasets, including PIE-SAR~\cite{zhang2024asanet}, YESeg-SAR~\cite{wei2024mgfnet}, FUSAR-Map~\cite{shi2021object}, DDHR-Korea~\cite{ren2022dual}, DDHR-SD~\cite{ren2022dual}, DDHR-XA~\cite{ren2022dual}, WHU-SAR~\cite{li2022mcanet} and OpenEarthMap-SAR~\cite{xia2025openearthmap}. These datasets represent various SAR sensing scenarios and target types.

% \noindent
% \textbf{Semantic segmentation.} We evaluate SegEarth-OV on 8 remote sensing semantic segmentation datasets including OpenEarthMap \cite{xia2023openearthmap}, LoveDA \cite{loveda}, iSAID \cite{waqas2019isaid}, Potsdam, Vaihingen\footnote{https://www.isprs.org/education/benchmarks/UrbanSemLab}, UAVid \cite{LYU2020108}, UDD5 \cite{chen2018large} and VDD \cite{cai2023vdd}. Among them, the first 5 datasets consist of mainly satellite images and the last 3 consist of UAV images. They contain custom foreground classes and a background class. Detailed descriptions of these datasets can be found in Appendix 7.1.
% % \ref{sec:supp_datasets_ss}.

% \noindent
% \textbf{Single-class extraction.} We select 4 building extraction datasets (\textit{i.e.}, WHU$^{Aerial}$ \cite{ji2018fully}, WHU$^{Sat.\mathrm{II}}$ \cite{ji2018fully}, Inria \cite{maggiori2017can}, and xBD \cite{gupta2019xbddatasetassessingbuilding}), 4 road extraction datasets (\textit{i.e.}, CHN6-CUG \cite{zhu2021global}, DeepGlobe \cite{demir2018deepglobe}, Massachusetts \cite{MnihThesis}, and SpaceNet \cite{van2018spacenet}), and 1 flood detection dataset (\textit{i.e.}, WBS-SI\footnote{https://www.kaggle.com/datasets/shirshmall/water-body-segmentation-in-satellite-images}) for the evaluation of single-class extraction. These datasets contain 1 foreground class (building, road or flood) and 1 background class. Detailed descriptions are in Appendix 7.2-7.4.
% % \ref{sec:supp_datasets_b}-\ref{sec:supp_datasets_f}. 

\noindent
\textbf{Training dataset for SimFeatUp.} SimFeatUp requires only raw image data for its initial training phase. To avoid any unfair comparisons or domain biases, we train SimFeatUp on a publicly available optical remote sensing image classification dataset, Million-AID \cite{Long2021DiRS}. We randomly selected 16k images from Million-AID for this purpose. This initial training is independent of the OVSS task itself, ensuring SimFeatUp's weights are universally applicable across different remote sensing modalities without requiring further fine-tuning.

\noindent
\textbf{Optical-SAR Paired Datasets for AlignEarth.} To train AlignEarth's SAR image encoder, we collect a large-scale paired optical-SAR dataset comprising 65k image pairs from various public sources. The datasets include: SpaceNet 6~\cite{shermeyer2020spacenet}, MSAW~\cite{zhang2025multi}, QXS-SAROPT~\cite{huang2021qxs}, SAR2Opt~\cite{zhao2022comparative}, DFC2023 Track1~\cite{persello20232023}, PIE-RGB-SAR~\cite{zhang2024asanet}, DFC2025 Track1~\cite{xia2025openearthmap}, WHU-OPT-SAR~\cite{li2022mcanet}, DDHR-SK~\cite{ren2022dual}, FUSAR-Map~\cite{shi2021object}, and YESeg-OPT-SAR~\cite{wei2024mgfnet}. These datasets offer diverse scenes, resolutions, and co-registration qualities, essential for robust cross-modal knowledge distillation.

\begin{table*}[t]
  \caption{Open-vocabulary semantic segmentation quantitative comparison on SAR remote sensing datasets. Evaluation metric: mIoU. \textcolor{tabred}{\textbf{Best}} and \textcolor{tabblue}{\textbf{second best}} performances are highlighted. ``Oracle'' is achieved by a fully supervised SegFormer~\cite{xie2021segformer} model using full training data, representing the upper bound.}
  \label{tab:table_main_sar}
  \centering
  \scalebox{0.85}{
  \begin{tabular}{@{}llcccccccc|c@{}}
    \toprule[1pt]
    Models & & OpenEarthMap-SAR & DDHR-Korea & DDHR-SD & DDHR-XA & FUSAR-Map & WHU-SAR & YESeg-SAR & PIE-SAR & Average \\
    \midrule[1pt]
    CLIP & {OpenAI} & 5.5 & 13.2 & 10.3 & 6.4 & 7.6 & 5.1 & 12.0 & 10.3 & 8.8 \\
    & {AlignEarth} & 7.0 & 21.9 & 13.0 & 12.5 & 11.5 & 5.2 & 12.2 & 15.0 & 12.3 \\
    \midrule[1pt]
    MaskCLIP & {OpenAI} & 6.7 & 13.0 & 13.1 & 7.8 & 9.8 & 5.6 & 14.0 & 12.8 & 10.4 \\
    & {AlignEarth} & 12.4 & 28.5 & 24.9 & 24.6 & 17.6 & 9.6 & 20.3 & 32.2 & 21.3 \\
    \midrule[1pt]
    SCLIP & {OpenAI} & 6.6 & 12.5 & 12.0 & 5.7 & 4.0 & 5.4 & 13.8 & 19.7 & 10.0 \\
    & {AlignEarth} & 18.6 & 42.5 & 31.4 & 38.7 & 23.0 & 12.9 & 26.2 & 41.6 & 29.4 \\
    \midrule[1pt]
    ClearCLIP & {OpenAI} & 7.5 & 9.8 & 8.5 & 4.2 & 15.7 & 4.7 & 14.8 & 20.6 & 10.7 \\
    & {AlignEarth} & \textcolor{tabblue}{\textbf{18.3}} & \textcolor{tabblue}{\textbf{44.6}} & \textcolor{tabblue}{\textbf{34.4}} & \textcolor{tabblue}{\textbf{35.9}} & \textcolor{tabblue}{\textbf{24.5}} & \textcolor{tabred}{\textbf{17.2}} & \textcolor{tabblue}{\textbf{25.6}} & \textcolor{tabblue}{\textbf{40.7}} & \textcolor{tabblue}{\textbf{30.2}} \\
    \midrule[1pt]
    SegEarth-OV & {OpenAI} & 6.3 & 16.5 & 15.1 & 7.3 & 9.7 & 5.1 & 13.0 & 12.1 & 10.6 \\
    & {AlignEarth} & \textcolor{tabred}{\textbf{19.3}} & \textcolor{tabred}{\textbf{48.6}} & \textcolor{tabred}{\textbf{36.6}} & \textcolor{tabred}{\textbf{44.5}} & \textcolor{tabred}{\textbf{28.2}} & \textcolor{tabblue}{\textbf{17.0}} & \textcolor{tabred}{\textbf{26.8}} & \textcolor{tabred}{\textbf{51.1}} & \textcolor{tabred}{\textbf{34.0}} \\
    \midrule[1pt]
    \textcolor{tablegray}{Oracle} & & \textcolor{tablegray}{35.1} & \textcolor{tablegray}{70.2} & \textcolor{tablegray}{63.6} & \textcolor{tablegray}{62.3} & \textcolor{tablegray}{37.3} & \textcolor{tablegray}{45.7} & \textcolor{tablegray}{43.8} & \textcolor{tablegray}{63.2} & \textcolor{tablegray}{52.7} \\
    \bottomrule[1pt]
  \end{tabular}}
\end{table*}

\begin{figure*}[t]
  \centering
%   \fbox{\rule{0pt}{2in} \rule{0.9\linewidth}{0pt}}
   \includegraphics[width=1.0\linewidth]{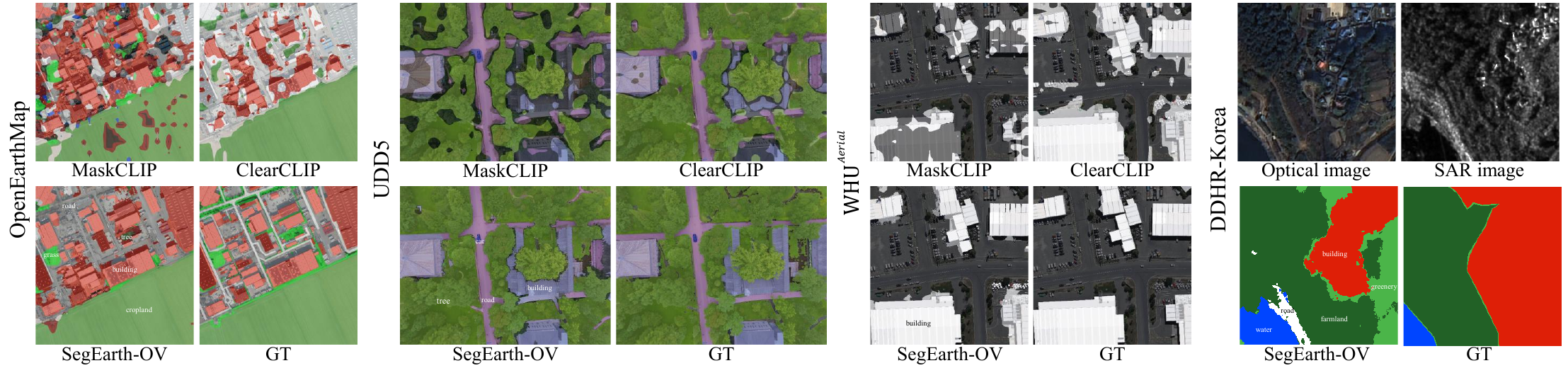}
   \caption{Qualitative comparison between different annotation-free OVSS methods on OpenEarthMap~\cite{xia2023openearthmap}, UDD5~\cite{chen2018large}, WHU$^{Aerial}$~\cite{ji2018fully} and DDHR-Korea~\cite{ren2022dual} datasets (best viewed digitally with zoom, especially for the edges of the object).}
   \label{fig:samples}
\end{figure*}

\subsection{Setup}

\noindent
\textbf{Implementation.} Our implementations are based on MMSegmentation \cite{mmseg2020} toolkit. If not specified, we use the original pretrained weights of CLIP (ViT-B/16) provided by OpenAI. For the text part, we use the \texttt{OpenAI ImageNet template} as input for the text encoder, \textit{e.g.}, ``a photo of a \{class name\}''. In addition, since the definition of certain classes may vary in some datasets, we use slight class rename tricks for all methods. For example, we rename ``clutter'' to ``background'' and ``building'' to \{``building'', ``house'', ``roof''\}, and the sub-class with the highest probability in \{\} will be the probability of that class. A detailed list of prompt class names for all datasets is provided in Table \ref{table_class_name}. For the image part, we resize input images with a long side of 448 and perform slide inference with a 224 $\times$ 224 window and 112 stride.

For SimFeatUp training, we randomly crop 224 $\times$ 224 image patches on the original image. We use two 4090 GPUs to train 1 epoch with batch size set to 8. We retain the multi-view consistency constraint in FeatUp, and random flipping, translation and zoom are applied. For AlignEarth training, the SAR image encoder $\mathcal{E}_{sar}$ is initialized from OpenAI CLIP's image encoder. We use the AdamW optimizer with a learning rate of 1e-4 and train for 20 epochs in a single A100 GPU. The batch size for AlignEarth training is 512. After training, the SAR encoder is frozen and integrated into SegEarth-OV. For the hyper-parameters mentioned, the value of $\gamma$ is set to 0.1, the temperature $\tau$ in the contrastive loss is initialized to 0.07, the region size $K$ of local distillation is set to $7$, and $\lambda$ is set to 0.3 for all datasets.

\noindent
\textbf{Evaluation.} We evaluate the semantic segmentation using the mean intersection over union (mIoU) metric. For single-class extraction, the IoU of the foreground class is used.

\noindent
\textbf{Baseline.} We incorporate common practices from natural image OVSS, which are also suitable for remote sensing scenes. These include removing the FFN and residual connection of the last Transformer block \cite{li2023clip}, \cite{lan2024clearclip}. In addition, the last self-attention is replaced by a modulated attention, \textit{i.e.}, the summation of \textit{q-q}, \textit{k-k} and \textit{v-v} as the weights of \textit{v}.

% \vspace{-1em}
% \begin{equation}
% \begin{aligned}
% \operatorname{M-SA}(\boldsymbol{q}, \boldsymbol{k}, \boldsymbol{v}) = \hspace{-1em} \sum_{\boldsymbol{i}\in \{\boldsymbol{q}, \boldsymbol{k}, \boldsymbol{v}\}}\hspace{-1em}\text{softmax}(\frac{\boldsymbol{i} \cdot \boldsymbol{i}^\mathsf{T}}{\sqrt{d}}) \cdot \boldsymbol{v}.
% \label{eq:msa}
% \end{aligned}
% \end{equation}

\subsection{Comparison with State-of-the-art}

Since the proposed SegEarth-OV is a annotation-free method and there is no previous OVSS method designed for remote sensing, we select 5 SOTA annotation-free OVSS models of natural images for comparison, including vanilla CLIP \cite{radford2021learning}, MaskCLIP \cite{zhou2022extract}, SCLIP \cite{wang2023sclip}, GEM \cite{bousselham2024grounding} and ClearCLIP \cite{lan2024clearclip}.

\subsubsection{Performance on Optical Images}

\noindent
\textbf{Semantic segmentation.} As listed in Table~\ref{table_main}, SegEarth-OV achieves the best performance on all 8 semantic segmentation datasets. SegEarth-OV achieves more than 40\% mIoU on 5 datasets and more than 50\% on the UDD5 dataset, which implies that the OVSS method is feasible in remote sensing scenarios. Compared to the previous method, SegEarth-OV achieves a performance gain of more than 5\% on 5 datasets and an average gain of 5.8\% on 8 datasets. On the iSAID dataset, the mIoU of SegEarth-OV is only 21.7\%, which is due to the fine-grained category delineation in this dataset, which covers 16 categories (see Table~\ref{table_class_name}). Compared to the conference version~\cite{li2025segearth}, we used more suitable category name prompts on the Potsdam and Vaihingen datasets, replacing the original ``low vegetation'' with ``grass'', a more common term that better matches the style of the text encoder when pre-trained on natural images. The ``Oracle'' results, achieved by a fully supervised SegFormer~\cite{xie2021segformer}, represent the upper bound of performance with full training data. Although our annotation-free method does not reach this ceiling, SegEarth-OV narrows the gap significantly.

\noindent
\textbf{Single-class extraction.} In the building extraction task, the increase delivered by SegEarth-OV is more significant, as listed in Table~\ref{table_main_br}. Considering that the ``building'' occupies a small area, we evaluate the setup for larger scale images, \textit{i.e.}, resizing the long side of the input image to 896. This setup significantly improves the IoU of Inria and xBD, which on the other hand supports our view that spatial detail preservation is essential for remote sensing OVSS. In the road extraction task, although SegEarth-OV achieves the best IoU, overall, the performance of all methods on the 4 datasets is unsatisfactory, with a best IoU of only 35.4\%. This may be due to two reasons: (1) the special shape of the road makes it difficult to be extracted in a annotation-free OVSS manner; (2) the labels of some data are generated based on OpenStreetMap\footnote{https://www.openstreetmap.org/} vector shapes with fixed widths attached, which are inherently imprecise. Again, the extraction of roads can generally benefit from larger size images. For the flood detection task, where ``water'' class features can be easily recognized, the IoU of SegEarth-OV is improved by 15.3\% over the previous best method, up to 60.2\%. Due to the small size of the original images in the WBS-SI dataset, resizing to a larger size does not result in a positive gain.

% One observation is that the performance achieved by OVSS is correlated with the spatial resolution, \textit{e.g.}, WHU$^{Aerial}$ and Intra have the highest resolution in the 4 datasets and their IoUs reach 49.2\% and 44.6\%.

\begin{table*}
  \caption{Quantitative comparison of vanilla CLIP and remote sensing CLIPs (ViT-B/32). Evaluation metric: mIoU.}
  \label{table_rsclip}
  \centering
  \scalebox{1.0}{
  \begin{tabular}{@{}llcccccccc|c@{}}
    \toprule[1pt]
    Models & & OpenEarthMap & LoveDA & iSAID & Potsdam & Vaihingen & UAVid$^{img}$ & UDD5 & VDD & Average \\
    \midrule[1pt]
    CLIP \cite{radford2021learning} & {\tiny ICML'21} & 25.7 & 27.2 & 16.2 & \textcolor{tabblue}{\textbf{40.0}} & \textcolor{tabred}{\textbf{25.1}} & \textcolor{tabblue}{\textbf{31.6}} & \textcolor{tabred}{\textbf{39.7}} & \textcolor{tabblue}{\textbf{39.1}} & 30.6 \\
    \midrule[1pt]
    RemoteCLIP \cite{liu2024remoteclip} & {\tiny TGRS'23} & 18.2 & \textcolor{tabred}{\textbf{37.8}} & \textcolor{tabblue}{\textbf{18.9}} & 21.9 & 22.9 & 16.1 & 27.1 & 28.1 & 23.9 \\
    GeoRSCLIP \cite{zhang2023rs5m} & {\tiny TGRS'24} & \textcolor{tabred}{\textbf{35.0}} & 30.8 & \textcolor{tabred}{\textbf{23.6}} & 38.0 & 22.3 & \textcolor{tabred}{\textbf{34.0}} & \textcolor{tabblue}{\textbf{39.1}} & \textcolor{tabred}{\textbf{40.5}} & \textcolor{tabred}{\textbf{32.9}} \\
    SkyCLIP \cite{wang2024skyscript} & {\tiny AAAI'24} & \textcolor{tabblue}{\textbf{28.6}} & \textcolor{tabblue}{\textbf{33.0}} & 15.3 & \textcolor{tabred}{\textbf{41.7}} & \textcolor{tabblue}{\textbf{24.1}} & \textcolor{tabblue}{\textbf{31.6}} & 38.2 & 35.8 & \textcolor{tabblue}{\textbf{31.0}} \\
    \bottomrule[1pt]
  \end{tabular}}
\end{table*}

\subsubsection{Performance on SAR Images}

For SAR image segmentation, we establish baselines by directly applying SOTA optical OVSS methods to SAR images and compare them with our AlignEarth-powered versions. Table~\ref{tab:table_main_sar} presents the quantitative results on 8 SAR datasets. The ``OpenAI'' involves directly applying the original CLIP to SAR images, while the ``AlignEarth'' uses our distilled SAR encoder. As expected, directly applying optical models to SAR images yields poor performance due to the significant modality gap. In stark contrast, our AlignEarth strategy brings about dramatic improvements across the board. For every baseline method (CLIP, MaskCLIP, SCLIP, ClearCLIP), integrating the AlignEarth's SAR encoder leads to substantial performance gains. Most importantly, our full SegEarth-OV framework with AlignEarth sets a new SOTA for SAR OVSS, achieving the best performance on 7 datasets. For instance, on the DDHR-Korea dataset, our method achieves 48.6\% mIoU, a massive improvement over the direct application baseline (16.5\%). On PIE-SAR, we achieve 51.1\% mIoU, demonstrating our framework's feasibility in practical SAR scenes. On some datasets, SegEarth-OV with AlignEarth performs close to the ``Oracle''. For instance, on PIE-SAR, our method reaches over 80\% of the supervised oracle's performance (63.2\%). These results unequivocally validate that AlignEarth successfully bridges the optical-SAR modality gap, and when combined with SimFeatUp and Global Bias Alleviation, provides a powerful and universal solution for multi-modal remote sensing OVSS.

\noindent
\textbf{Qualitative results.} We present qualitative results for MaskCLIP, ClearCLIP, and SegEarth-OV in Fig.~\ref{fig:samples}. Some observations are summarized as follows:  (1) There are some incorrect category predictions in MaskCLIP, \textit{e.g.}, \textcolor[RGB]{0, 69, 255}{water} on the road and \textcolor[RGB]{128, 0, 0}{bareland} on the \textcolor[RGB]{75, 181, 73}{cropland}. (2) ClearCLIP can generate correct category predictions, but lacks precise localization capability, with distorted target shapes and ill-fitting boundaries of the prediction mask. (3) SegEarth-OV is capable of generating more fine-grained masks that fit the target edges and maintain correct category discrimination. (4) For SAR images, although our predictions differ significantly from the ground truth, it is clear that our predictions have more detailed mask divisions. This demonstrates the potential of the OVSS method.

% 对于SAR图像，虽然我们的预测比起ground truth有显著的差异，但明显的是，我们的预测中具有更细节的掩码划分。
% More visualizations can be found in Appendix Fig. 8-10.
% \cref{fig:OEM}-\ref{fig:WHU}.

\subsection{Ablation Study and Analysis}

\noindent
\textbf{Plug and play.} SimFeatUp and Global Bias Alleviation can be attached to other OVSS methods as plug-and-play modules. As listed in Table~\ref{table_compare}, a revealing observation is that on both the OpenEarthMap and WHU$^{Aerial}$ datasets, as the base capability of the model improves (from MaskCLIP to ClearCLIP), the increases delivered by our method also improve (\textbf{{\color{green!50!black}$\uparrow$\scriptsize{3.3}}}, \textbf{{\color{green!50!black}$\uparrow$\scriptsize{5.1}}}, \textbf{{\color{green!50!black}$\uparrow$\scriptsize{8.1}}} on OpenEarthMap, \textbf{{\color{green!50!black}$\uparrow$\scriptsize{5.6}}}, \textbf{{\color{green!50!black}$\uparrow$\scriptsize{6.1}}}, \textbf{{\color{green!50!black}$\uparrow$\scriptsize{14.5}}} on WHU$^{Aerial}$). This suggests that our method has the potential to improve localization and discrimination for stronger models. 

% In addition, ``ClearCLIP + ours'' outperforms SegEarth-OV on WHU$^{Aerial}$ and WBS-SI, which suggests that the modulated attention we use is not optimal in some cases, and exploring how to design a better self-attention in annotation-free OVSS is meaningful. 

% In addition, SimFeatUp, as a plug-and-play module with only $<$ 0.3M parameters, is also effective on natural images, as listed in Appendix \cref{table_natural}.

\begin{table}
  \caption{The proposed method is a plug-and-play module. ``GBA'' indicates Global Bias Alleviation.}
  \label{table_compare}
  \centering
  \scalebox{1}{
  \begin{tabular}{@{}lccc@{}}
    \toprule[1pt]
    Methods & OpenEarthMap & {\color{red!50!black}WHU$^{Aerial}$} & {\color{blue!50!black}WBS-SI} \\
    \midrule[1pt]
    MaskCLIP & 25.1 & 29.8 & 39.8 \\
    \rowcolor{gray!20}
    + SimFeatUp\&GBA & 28.4\textbf{{\color{green!50!black}$\uparrow$\scriptsize{3.3}}} & 35.4\textbf{{\color{green!50!black}$\uparrow$\scriptsize{5.6}}} & 48.8\textbf{{\color{green!50!black}$\uparrow$\scriptsize{9.0}}} \\
    SCLIP & 29.3 & 33.4 & 32.1 \\
    \rowcolor{gray!20}
    + SimFeatUp\&GBA & 34.4\textbf{{\color{green!50!black}$\uparrow$\scriptsize{5.1}}} & 39.5\textbf{{\color{green!50!black}$\uparrow$\scriptsize{6.1}}} & 53.4\textbf{{\color{green!50!black}$\uparrow$\scriptsize{21.3}}} \\
    ClearCLIP & 31.0 & 36.6 & 44.9 \\
    \rowcolor{gray!20}
   + SimFeatUp\&GBA & 39.1\textbf{{\color{green!50!black}$\uparrow$\scriptsize{8.1}}} & 51.1\textbf{{\color{green!50!black}$\uparrow$\scriptsize{14.5}}} & 60.4\textbf{{\color{green!50!black}$\uparrow$\scriptsize{15.5}}} \\
    \bottomrule[1pt]
  \end{tabular}}
\end{table}

\begin{table}
  \caption{Detailed ablation results for each component. ``X''$\uparrow$ indicates upsampling earlier stage features, \textit{i.e.} \cref{eq:which_fea}. ``+ RS Data'' indicates using  Million-AID \cite{Long2021DiRS} to train the upsampler, before using the images in COCO-Stuff \cite{caesar2018coco}.}
  \label{table_ablation}
  \centering
  \scalebox{1}{
  \begin{tabular}{@{}l|c|c|c@{}}
    \toprule[1pt]
    & OpenEarthMap & {\color{red!50!black}WHU$^{Sat.\mathrm{II}}$} & {\color{blue!50!black}WBS-SI} \\
    \midrule[1pt]
    \textit{Baseline} & 32.4 & 22.7 & 46.9 \\
    FeatUp (CLIP)~\cite{fu2024featup} & 33.9 & 20.2 & 39.6 \\
    FeatUp (MaskCLIP)~\cite{fu2024featup} & 33.8 & 25.2 & 45.9 \\
    ``X''$\uparrow$ & 34.6 & 26.0 & 54.2 \\
    + RS Data & 36.0 \textbf{{\color{green!50!black}$\uparrow$\scriptsize{1.4}}} & 26.2 & 56.4 \\
    + JBU\_One & 36.3 \textbf{{\color{green!50!black}$\uparrow$\scriptsize{0.3}}} & 26.0 & 57.1 \\
    + Rec. Image& 37.6 \textbf{{\color{green!50!black}$\uparrow$\scriptsize{1.3}}} & 26.4 & 58.7 \\
    + Alleviate Global Bias & 39.3 \textbf{{\color{green!50!black}$\uparrow$\scriptsize{1.7}}} &  27.9 & 59.5 \\
    + Large Kernel & 40.3 \textbf{{\color{green!50!black}$\uparrow$\scriptsize{1.0}}} & 28.4 & 60.2 \\
    \bottomrule[1pt]
  \end{tabular}}
\end{table}

\begin{table}
  \caption{OVSS quantitative comparison on natural image datasets. The basic results are cited from \cite{lan2024clearclip}.}
  \label{table_natural}
  \centering
  \scalebox{0.95}{
  \begin{tabular}{@{}lccc|c@{}}
    \toprule[1pt]
    Methods & Context59 \cite{mottaghi2014role} & Stuff \cite{caesar2018coco} & Cityscapes \cite{cordts2016cityscapes} & Average \\
    \midrule[1pt]
    TCL \cite{cha2023learning} & 30.3 & 19.6 & 23.1 & 24.3 \\
    Reco \cite{shin2022reco} & 22.3 & 14.8 & 21.1 & 19.4 \\
    \midrule[1pt]
    MaskCLIP & 26.4 & 16.4 & 12.6 & 18.5 \\
    \rowcolor{gray!20}
    + SimFeatUp & 28.7 & 18.0 & 25.8 & 24.2 \textbf{{\color{green!50!black}$\uparrow$\scriptsize{5.7}}} \\
    SCLIP & 33.0 & 21.1 & 29.1 & 27.7 \\
    \rowcolor{gray!20}
    + SimFeatUp & 34.1 & 22.0 & 30.5 & 28.9 \textbf{{\color{green!50!black}$\uparrow$\scriptsize{1.2}}}\\
    ClearCLIP & 35.9 & 23.9 & 30.0 & 29.9 \\
    \rowcolor{gray!20}
    + SimFeatUp & 37.5 & 25.1 & 30.7 & 31.1 \textbf{{\color{green!50!black}$\uparrow$\scriptsize{1.2}}}\\
    \bottomrule[1pt]
  \end{tabular}}
\end{table}

\begin{table*}[t]
  \caption{Quantitative comparison of various CLIPs on SAR segmentation task. Evaluation metric: mIoU. \textcolor{tabred}{\textbf{Best}} and \textcolor{tabblue}{\textbf{second best}} performances are highlighted.}
  \label{table_sar_vs_rsclip}
  \centering
  \scalebox{0.9}{
  \begin{tabular}{@{}lcccccccc|c@{}}
    \toprule[1pt]
    Methods& OpenEarthMap-SAR & DDHR-Korea & DDHR-SD & DDHR-XA & FUSAR-Map & WHU-SAR & YESeg-SAR & PIE-SAR & Average \\
    \midrule[1pt]
    CLIP~\cite{radford2021learning} & 7.5 & 9.8 & 8.5 & 4.2 & 15.7 & 4.7 & 14.8 & 20.6 & 10.7 \\
    \midrule[1pt]
    OpenCLIP~\cite{cherti2023reproducible} & 4.9 & 13.3 & 14.7 & 8.3 & 11.5 & 5.6 & 14.0 & 14.5 & 10.9 \\
    MetaCLIP~\cite{xu2023demystifying} & 6.1 & 12.0 & 12.2 & 9.8 & 7.8 & 6.1 & 14.1 & 16.0 & 10.5 \\
    ALIP~\cite{yang2023alip} & 5.0 & 20.8 & 16.1 & 9.0 & 10.8 & 5.3 & 12.8 & 15.7 & 12.0 \\
    SkyCLIP~\cite{wang2024skyscript} & 8.2 & 20.5 & 22.6 & 8.9 & \textcolor{tabblue}{\textbf{12.0}} & 9.3 & 15.7 & 26.2 & \textcolor{tabblue}{\textbf{15.4}} \\
    GeoRSCLIP~\cite{zhang2023rs5m} & \textcolor{tabblue}{\textbf{10.2}} & 17.5 & 20.8 & 5.9 & 4.8 & \textcolor{tabblue}{\textbf{10.9}} & \textcolor{tabblue}{\textbf{18.7}} & \textcolor{tabblue}{\textbf{32.1}} & 15.1 \\
    RemoteCLIP~\cite{liu2024remoteclip} & 4.9 & \textcolor{tabblue}{\textbf{22.6}} & \textcolor{tabblue}{\textbf{22.8}} & \textcolor{tabblue}{\textbf{10.0}} & 10.2 & 3.7 & 7.1 & 7.5 & 11.1 \\
    CLIPSelf~\cite{wuclipself} & 1.7 & 1.4 & 4.2 & 1.8 & 9.5 & 3.2 & 9.9 & 3.4 & 4.4 \\
    \midrule[1pt]
    AlignEarth & \textcolor{tabred}{\textbf{18.3}} & \textcolor{tabred}{\textbf{44.6}} & \textcolor{tabred}{\textbf{34.4}} & \textcolor{tabred}{\textbf{35.9}} & \textcolor{tabred}{\textbf{24.5}} & \textcolor{tabred}{\textbf{17.2}} & \textcolor{tabred}{\textbf{25.6}} & \textcolor{tabred}{\textbf{40.7}} & \textcolor{tabred}{\textbf{30.2}} \\
    \bottomrule[1pt]
  \end{tabular}}
\end{table*}

\noindent
\textbf{Ablation study.} To assess each of the proposed components, we perform a detailed ablation analysis, as listed in Table~\ref{table_ablation}. FeatUp (CLIP) denotes the original FeatUp upsampler~\cite{fu2024featup}, which provides a 1.5\% improvement on OpenEarthMap but decreases the performance on WHU$^{Sat.\mathrm{II}}$ and WBS-SI. FeatUp (MaskCLIP) denotes using $\boldsymbol{v}$ of self-attention as the upsampled feature, which somewhat mitigates the possible negative effects of FeatUp (CLIP). In SimFeatUp, the input feature $X$ of the last block is used to upsample, which presents a significant improvement in all 3 datasets. A substantial improvement is also delivered after replacing the training material for the upsampler from natural images to remote sensing images. ``JBU\_One'' reduces the parameters by nearly $4\times$ while delivering a slight IoU gain (only $<$ 0.3M parameters). The introduction of $\operatorname{CRN}$ with image reconstruction loss brings 1.7\%, 0.4\%, and 1.6\% improvement on 3 datasets, respectively. Note that the $\operatorname{CRN}$ only participates in SimFeatUp's training and is discarded during inference. Global bias alleviation shows significant improvement for all 3 datasets, with an average 1.3\% improvement. Finally, expanding the upsampling kernel to $11\times 11$ also exhibits consistent improvement across all datasets.

\noindent
\textbf{Results on natural images.} We evaluate SimFeatUp as an external unit on 3 natural image datasets: PASCAL Context59 \cite{mottaghi2014role}, COCOStuff \cite{caesar2018coco} and Cityscapes \cite{cordts2016cityscapes}. As listed in Table~\ref{table_natural}, after upsampling the visual features of MaskCLIP, SCLIP, and ClearCLIP using SimFeatUp, their mIoUs are improved by 5.7\%, 1.2\%, and 1.2\%, respectively. This reveals the potential of our method to inspire general vision.

\noindent
\textbf{Remote sensing CLIPs for OVSS.} We evaluate the performance of remote sensing CLIPs on OVSS, including RemoteCLIP \cite{liu2024remoteclip}, GeoRSCLIP \cite{zhang2023rs5m}, and SkyCLIP \cite{wang2024skyscript}, which are trained on 0.8M, 5M, and 2.6M remote sensing data, respectively, without changing the model structure of CLIP. Since these works do not provide the ViT-B/16, we uniformly use ViT-B/32. Hence, we repeat the JBU operation 5 times in SimFeatUp. For fair comparison, we train the respective SimFeatUp for each model. As listed in Table~\ref{table_rsclip}, RemoteCLIP performs suboptimally to vanilla CLIP, which indicates that a small amount of domain data diminishes the model's transfer capability. GeoRSCLIP achieves the best performance against SkyCLIP, which suggests that domain VLMs can benefit from more diverse domain-specific data. Moreover, the OVSS task effectively reflects the model's discrimination and localization capabilities, and can serve as an evaluation metric for remote sensing VLMs.

\noindent
\textbf{Analysis of AlignEarth Strategy.} 
To comprehensively evaluate the AlignEarth strategy, we compare its performance against existing CLIPs on the SAR OVSS task. As listed in Table~\ref{table_sar_vs_rsclip}, we categorize the competing models into 3 groups: general models (CLIP~\cite{radford2021learning}, OpenCLIP~\cite{cherti2023reproducible}, MetaCLIP~\cite{xu2023demystifying}, \textit{etc}.), domain-specific remote sensing models (SkyCLIP~\cite{wang2024skyscript}, GeoRSCLIP~\cite{zhang2023rs5m}, RemoteCLIP~\cite{liu2024remoteclip}), and segmentation-specific models (CLIPSelf~\cite{wuclipself}). All models are evaluated using the same OVSS logic (ClearCLIP) for a fair comparison. Our AlignEarth model achieves an average mIoU of 30.2\%, dramatically outperforming all other CLIPs. The best domain-specific model, SkyCLIP, only reaches an average mIoU of 15.4\%, while general models perform even worse.  This large performance gap highlights that simply pre-training on vast amounts of in-domain optical remote sensing data is insufficient to bridge the modality gap between optical and SAR image. The unique physics of SAR imaging requires a more targeted adaptation. CLIPSelf, although designed to be more suitable for dense prediction, suffers from a loss of generalization ability after being tuned on natural image segmentation datasets (COCO~\cite{lin2014microsoft}).

% CLIPSelf，虽然被设计更适合密集预测，但在自然图像分割数据集上的微调导致其泛化能力的损失。

\section{Conclusion}
\label{sec:conclusion}

In this paper, we present SegEarth-OV, a annotation-free OVSS method for remote sensing images. The design of SegEarth-OV was motivated by the observation that OVSS methods currently used for natural images do not perform well on remote sensing images. The two key insights of SegEarth-OV, \textit{i.e.}, SimFeatUp and Global Bias Alleviation, exhibit consistent improvements on 17 remote sensing datasets spanning the tasks of semantic segmentation, building extraction, road extraction, and flood detection, well beyond the previous state-of-the-art methods. Furthermore, to extend SegEarth-OV's versatility to other challenging remote sensing modalities like SAR image, we propose AlignEarth. This cross-modal knowledge distillation strategy efficiently transfers rich semantic understanding from optical VLMs to a dedicated SAR image encoder, enabling SegEarth-OV to perform annotation-free OVSS on SAR images. Crucially, as the first comprehensive exploration of an annotation-free OVSS solution in remote sensing, this work unequivocally demonstrates the feasibility and strong potential of applying open-vocabulary concepts to Earth perception tasks. We anticipate that SegEarth-OV will inspire the development of more advanced OVSS methods and more capable remote sensing VLMs, ultimately opening up exciting new possibilities for the broader Earth vision community.

\bibliographystyle{IEEEtran}
\bibliography{ref.bib}

\end{document}